\pdfoutput=1

\documentclass[11pt]{article}

\usepackage[final]{acl}

\usepackage{times}
\usepackage{latexsym}
\usepackage{todonotes}
\usepackage[T1]{fontenc}

\usepackage[utf8]{inputenc}

\usepackage{microtype}

\usepackage{inconsolata}

\usepackage{graphicx}
\usepackage{subcaption}

\usepackage{booktabs} 
\usepackage[table,xcdraw]{xcolor} 
\usepackage{adjustbox}
\usepackage{todonotes}
\usepackage{amsmath}
\usepackage{amssymb}
\usepackage{multirow}
\usepackage{fontawesome5}

\usepackage[breakable]{tcolorbox}
\newtcolorbox{promptbox}[1][]{
  colback=gray!5, colframe=gray!50,
  fonttitle=\bfseries, breakable,
  before skip=6pt, after skip=6pt,
  fontupper=\small\ttfamily\setlength{\parindent}{0pt}\setlength{\parskip}{4pt},
  #1
}
\newtcolorbox{promptbox-delta}[1][]{
  colback=gray!5, colframe=gray!50,
  fonttitle=\bfseries, breakable,
  before skip=6pt, after skip=6pt,
  fontupper=\small\ttfamily\setlength{\parindent}{0pt}\setlength{\parskip}{4pt},
  #1
}
\newtcolorbox{samplebox}[1][]{
  colback=gray!5, colframe=gray!50,
  fonttitle=\bfseries, breakable,
  before skip=6pt, after skip=6pt,
  fontupper=\small\ttfamily\setlength{\parindent}{0pt}\setlength{\parskip}{4pt},
  #1
}

\usepackage{caption}

\usepackage{tabularx}
\usepackage{placeins}
\usepackage{booktabs}

\usepackage{enumitem}

\usepackage{tikz}
\usetikzlibrary{positioning}

\definecolor{iconcolor}{gray}{0}

\newcommand{\hF}{{\small\faTimesCircle}}
\newcommand{\hS}{{\small\faCheckCircle}}
\newcommand{\hT}{{\small\faClock}}

\NewDocumentCommand{\hMapTable}{o}{%
  \textcolor{iconcolor}{%
    \raisebox{-1em}{%
      \begin{tikzpicture}[baseline=(big.base)]
        \node[inner sep=0pt] (big) {\LARGE\faIcon{map-marked-alt}};
        \IfValueT{#1}{%
          \fill[white] (big.south east) circle (1.3ex);
          \node[inner sep=0pt] at (big.south east) {#1};
        }%
      \end{tikzpicture}%
    }%
  }%
}

\NewDocumentCommand{\hCivilianTable}{o}{%
  \textcolor{iconcolor}{%
    \raisebox{-1em}{%
      \begin{tikzpicture}[baseline=(big.base)]
        \node[inner sep=0pt] (big) {\LARGE\faUsers[regular]};
        \IfValueT{#1}{%
          \fill[white] (big.south east) circle (1.3ex);
          \node[inner sep=0pt] at (big.south east) {#1};
        }%
      \end{tikzpicture}%
    }%
  }%
}

\newcommand{\hMap}{\textcolor{iconcolor}{\small\faIcon{map-marked-alt}}}
\newcommand{\hCivilian}{\textcolor{iconcolor}{\small\faUsers[regular]}}

\newcommand{\hStaticTable}{\textcolor{iconcolor}{\raisebox{-1.3em}{\LARGE\faStopCircle}}}
\newcommand{\hMovingTable}{\textcolor{iconcolor}{\raisebox{-1em}{\LARGE\faPlayCircle}}}

\newcommand{\hStatic}{\textcolor{iconcolor}{\small\faStopCircle}}
\newcommand{\hMoving}{\textcolor{iconcolor}{\small\faPlayCircle}}

\title{Guide Me Out: A Framework to Benchmark VLM Operators Communication in Crisis Scenarios}

\author{
 \textbf{Giacomo Gonella\textsuperscript{1, 2}},
 \textbf{Stefano Menini\textsuperscript{1}},
 \textbf{Marco Guerini\textsuperscript{1}}
\\
\textsuperscript{1}Fondazione Bruno Kessler, Italy,
\\
\textsuperscript{2}University of Trento, Italy
\\
 \texttt{\{ggonella, menini, guerini\}@fbk.eu}
}

\begin{document}
\maketitle

\begin{abstract}
Effective crisis response requires spatially grounded communication that bridges linguistic guidance of civilians with the physical environment, accounting for structural bottlenecks, evolving threats, and agent-specific contexts. Yet, current NLP research in crisis communication remains mainly limited to static, text-only classification settings, overlooking the critical communicative role of AI operators in dynamic, embodied scenarios. We address this gap with a novel benchmarking framework for evaluating Vision-Language Models (VLMs) tasked with guiding civilian agents through simulated evacuations. We test two communication strategies (narrowcast vs. broadcast), two environment representations (visual vs. graph-based), and two threat behaviors (static vs. moving) across nine maps of varying structural complexity. Our results show that Narrowcast consistently reduces civilian Fail rates compared to Broadcast across all difficulty levels. Guidance quality depends heavily on how the VLM operator represents the world: the visual modality drives performance, while adding an adjacency graph is model-dependent and often harmful. Moving threats raise Fail rates across all conditions as communication must continuously adapt over time. Together, these findings show that deploying VLMs as AI operators in evacuation scenarios remains a non-trivial challenge, where the choice of communication strategy and input representation can directly determine the success or failure of the intervention.
\end{abstract}

\section{Introduction}

During emergency situations, operators must frequently coordinate the simultaneous routing of multiple civilians to safety. Civilians are dispersed and face heterogeneous, rapidly changing conditions, so a single broadcast message may be insufficient. Communication literature has shown that targeted messaging improves evacuation decisions \cite{Gao2021targetedwarning} and individual decision-making \cite{cao2017towards}, while excessive or poorly targeted messages lead to warning fatigue \cite{sutton2025overalerting}. Complementarily, evacuation research focuses on intervention efficiency \cite{Huang2023unityevaquation, dang2025llmfireevaquation}, relying on simulation to model civilian behavior but overlooking communication aspects. 

In this context, NLP can play a crucial role, yet, most research has focused on classifying social media posts in static, text-only settings. Real-world crises, however, unfold in physical environments with evolving dynamics, where operators must provide situated, time-critical guidance to help civilians navigate through space. This motivates investigating whether narrowcasting (individualized, continuous guidance) better supports individuals than broadcasting (general, periodic messages) and under which conditions.

\begin{figure}[t]
        \centering
        \includegraphics[width=\linewidth]{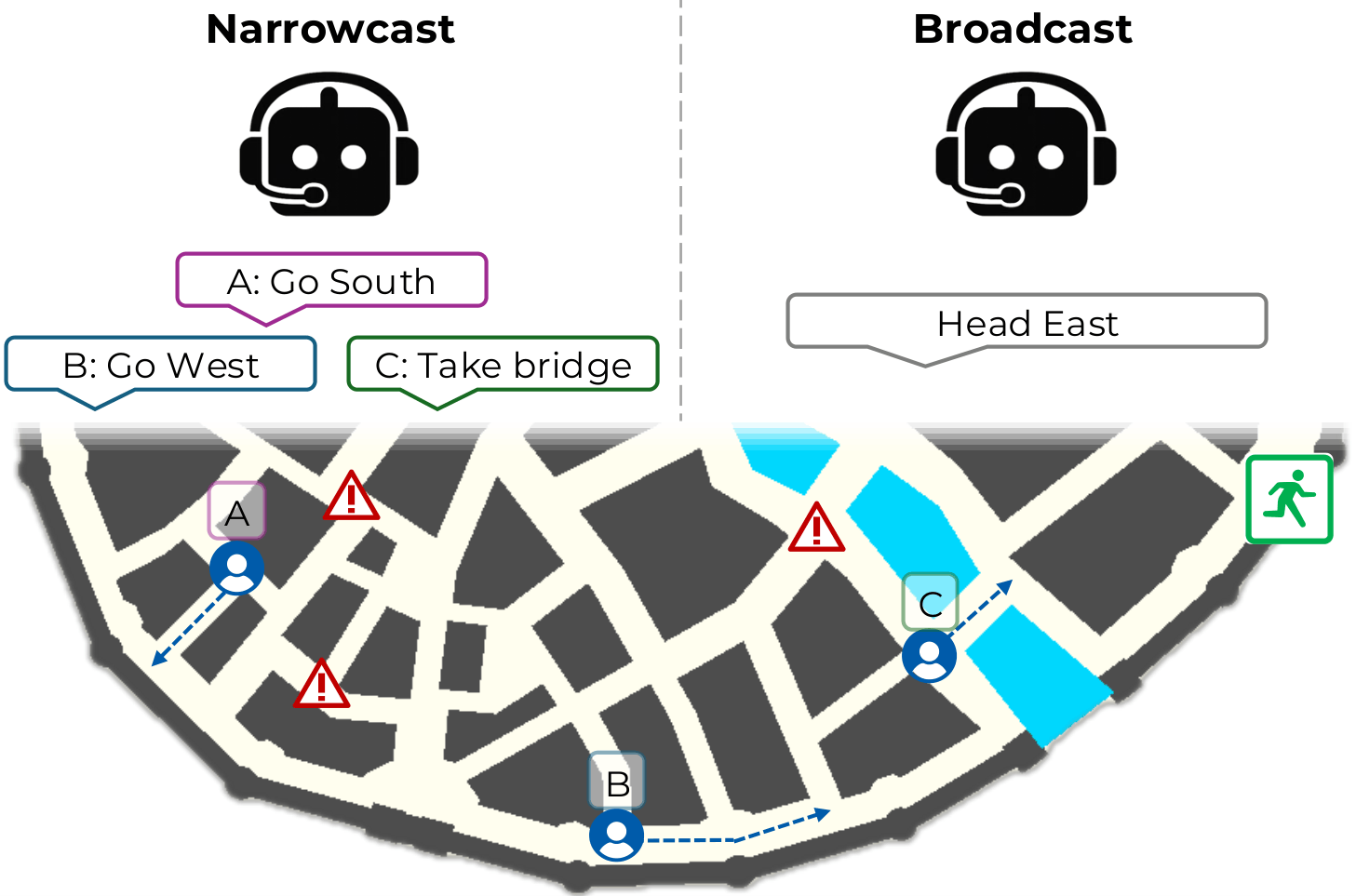}
        \caption{Two communication strategies on the same task. Narrowcast issues a tailored message to each civilian; broadcast issues one shared message to all.}
        \label{fig:teaser}
\end{figure}

Capturing the complexity of scenarios through a purely textual representation of the environment is difficult, motivating the use of a Visual Language Model (VLM) as operators. However, reasoning over visual relationships and spatially grounded environments remains non-trivial in current VLMs \cite{thrush2022winoground}, leaving unclear which input forms most effectively guide model behavior.

To address this gap, we introduce a simulation-based benchmarking framework, recreating scenarios that cannot be staged in reality, in which a VLM operator must guide civilians through nine urban maps of increasing structural difficulty. We use it to investigate three research questions:

\begin{tcolorbox}[colback=gray!5, colframe=gray!50, boxsep=1pt, left=3pt, right=3pt]
\textbf{RQ1.} How does narrowcast compare to broadcast for evacuation effectiveness?\\
\textbf{RQ2.} How do variations in threat dynamics impact effectiveness of narrowcast and broadcast?\\
\textbf{RQ3.} Which environment representation is more important for operator guidance?
\end{tcolorbox}
Our contribution is a simulation-based benchmarking framework for VLM-guided evacuation that varies communication strategy (narrowcast vs.~broadcast), threat dynamics (static vs.~moving), and environment representation (visual vs.~structured) across maps of increasing difficulties.\footnote{Framework code available at: \url{https://github.com/LanD-FBK/guide-me-out}} Results show that Narrowcast consistently reduces Fail rates, moving threats are more challenging and visual input is essential, while graph structure helps only in limited cases.

\section{Related Work}

\paragraph{Crisis-related NLP research.} has largely relied on social media data. The diffusion of social networks has allowed the creation of datasets from user-generated content during crisis events like floods or tornadoes \cite{olteanu2014crisislex, imran2016twitter}. These datasets, typically tweet-based with human annotations, support binary or multi-label classification tasks over dimensions such as informativeness, crisis detection, or crisis type recognition. These tasks are addressed either by training or fine-tuning deep learning models \cite{alam2021crisisbench, liu2021crisisbert} or using LLMs for zero-shot classification \cite{yin2024crisissense}. 

Beyond classification, crisis communication research also addresses operator–civilian interactions. LLM-based systems have been proposed to assist operators during emergency calls by extracting relevant information in real time \cite{otal2024llmassisted}, and to support operator training in high-stakes conversations such as interrogations or negotiations \cite{violakis2025leveraging}. Another line of research focused on NLG, producing warning messages across emergency scenarios \citep{gonella2026crisitext}, though still in purely textual and broadcast settings.

\paragraph{Simulations} have long been used to run experiments across various fields \cite{winsberg2013computer, gilbert2005simulation}, particularly when real-world situations are difficult, costly, or impossible to reproduce \cite{balci1997verification}. They also serve as data collection processes with full experimental control, providing a proxy of reality for training AI models \cite{tobin2017domain}.

In NLP, simulations have been used to obtain social interaction data from LLM-based agents \cite{park2023generativeagents}. Agents can be assigned different individual profiles, enabling the study of human-like behavior. In robotics, simulated environments are used to train agents before testing in the real world \cite{savva2019habitatplatformembodiedai}. A similar approach is found in evacuation studies, an adjacent field of crisis management, where simulations improve evacuation efficiency through physical environment modeling \cite{Huang2023unityevaquation} and LLM-based human behavior simulation \cite{dang2025llmfireevaquation}.

Vision-and-language navigation (VLN) involves an agent following natural language instructions in a physical environment. Some works focus on navigation tasks \cite{zhou2024navgpt}, others on interactive dialogue to guide a human through an environment \cite{de2018talk}. Recent work has investigated how the environment should be represented for LM-based navigation, comparing visual and structured textual encodings for navigation instruction evaluation \citep{shami2026groke} and vision-language navigation \citep{liu2026tagavlm}.

\section{Benchmarking Framework}

\begin{figure*}[ht]
    \centering
    \begin{minipage}{0.95\textwidth}
        \centering
        \begin{subfigure}[t]{0.32\linewidth}
            \centering
            \includegraphics[width=\linewidth]{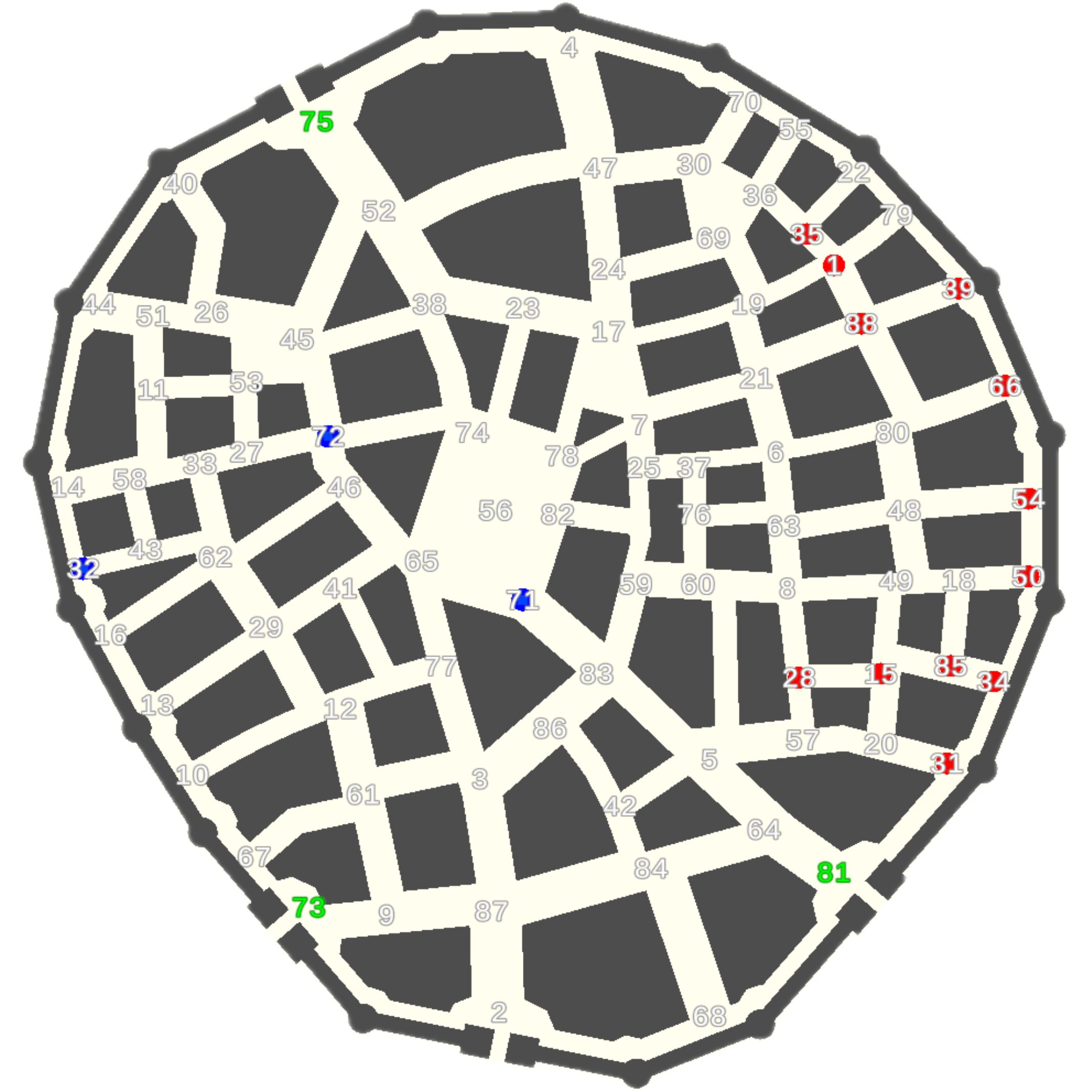}
        \end{subfigure}
        \hfill
        \begin{subfigure}[t]{0.32\linewidth}
            \centering
            \includegraphics[width=\linewidth]{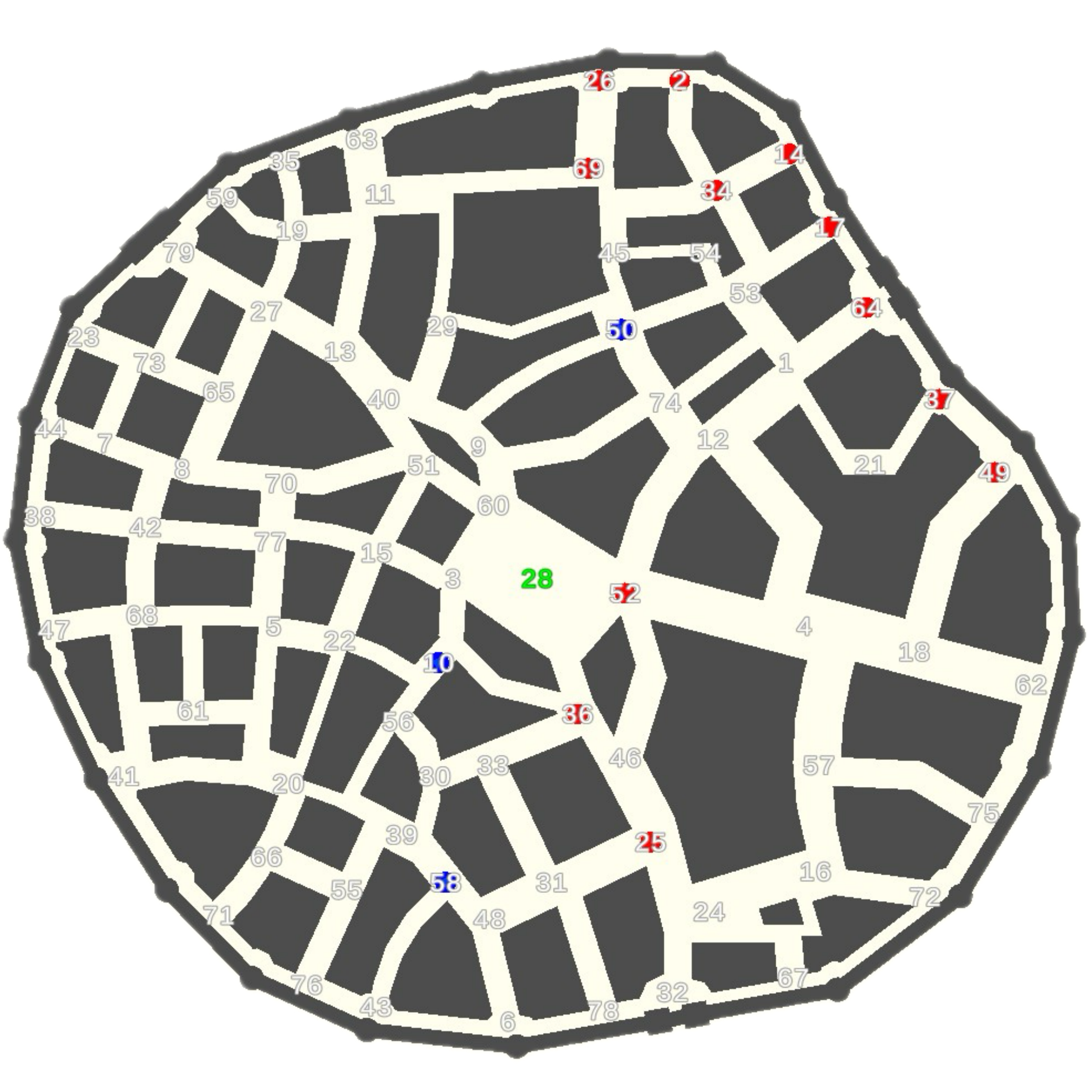}
        \end{subfigure}
        \hfill
        \begin{subfigure}[t]{0.32\linewidth}
            \centering
            \includegraphics[width=\linewidth]{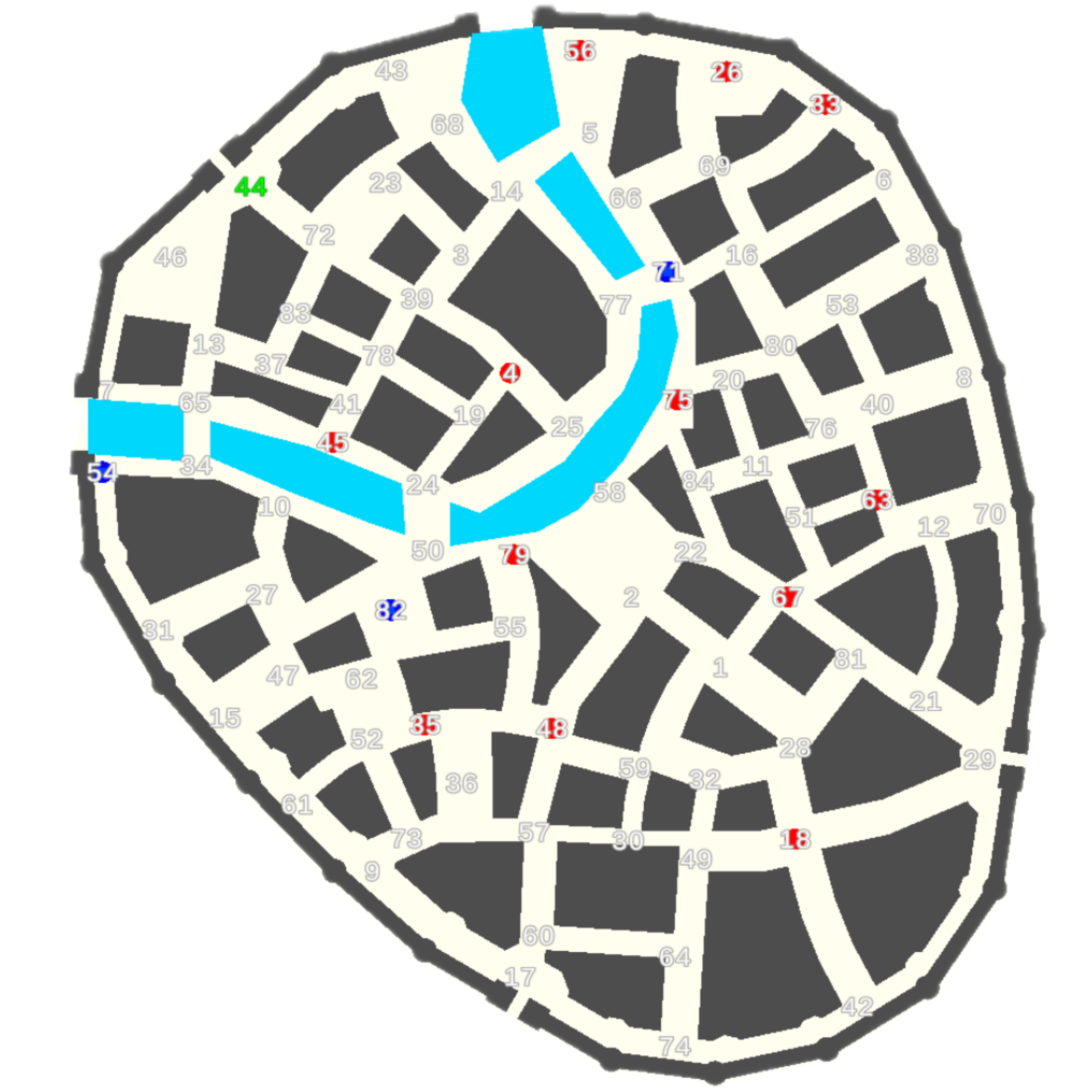}
        \end{subfigure}
    \end{minipage}
\caption{Easy (left), Medium (center), and Hard (right) map examples. Blue blocks represent civilians, red blocks threats, and green blocks safe exits.}
  \label{fig:three-maps-example}
\end{figure*}

\begin{figure}[ht]
    \centering
    \includegraphics[width=\linewidth]{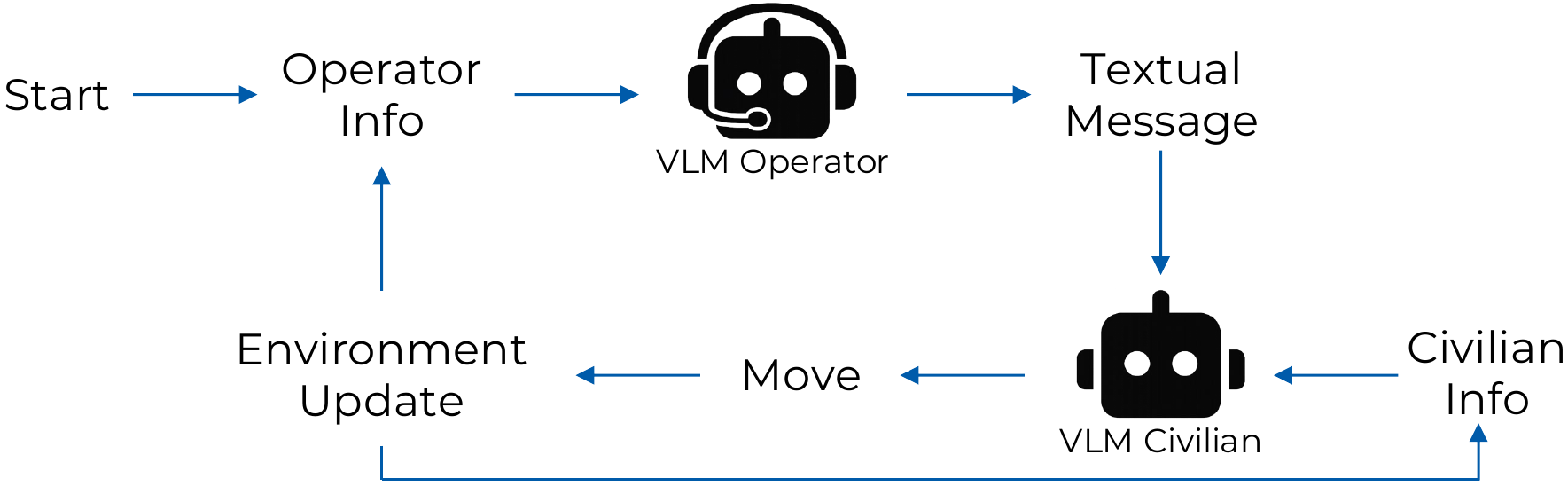}
    \caption{Simulation Turn Pipeline.}
    \label{fig:pipeline}
\end{figure}

To study the possible uses of AI for emergency communication in realistic crisis scenarios, we introduce a dynamic, spatially grounded benchmark. The benchmark is realized as a multi-agent simulation set in urban maps of increasing structural difficulty, where civilians and threats move while the operator, observing the scene, directs civilians toward a safe exit. Both operator and civilians are implemented as VLM agents, giving them visual perception and language understanding. Crucially, the operator holds a broader view of the environment than any civilian, and the benchmark is designed precisely to probe how this asymmetry can be handled through communication. Our simulations consist of a collection of \textit{episodes} and each episode proceeds in discrete turns: the operator observes the current state and issues guidance, civilians act on the messages they receive, and the cycle repeats until the scenario concludes. Figure~\ref{fig:pipeline} shows this mechanism and the following sections describe each element of the simulation in details.

\subsection{Environment}
\label{subsec:environment}

The simulated environments model urban areas with obstacles and navigable paths. Agents move on a graph of \textit{waypoints}: nodes are connected along the main routes, and a subset of waypoints is designated as \textit{exits}, the evacuation goals.\footnote{More details on map construction and waypoint naming are reported in Appendix~\ref{app:implementation}.} The simulation progresses in discrete time steps (\textit{turns}): at each turn, agents move to an adjacent waypoint. Each map contains (a) civilians to be guided to an exit and (b) threats to be avoided. Threats are stationary or mobile, with the latter performing a random walk (full dynamics in Section~\ref{sec:experimental-design}).

Informed by established evacuation literature, we construct scenarios of increasing difficulty by varying the geometries across nine maps, which are stratified into three distinct tiers (an example of each in Figure~\ref{fig:three-maps-example}, all maps in Figure~\ref{fig:all-maps} in the appendix). \textit{Easy} maps feature multiple exits and open paths. \textit{Medium} maps reduce exits to a single central one, forcing convergent evacuation paths for civilians \cite{shrahily2025comparative}. \textit{Hard} maps introduce a river crossable only through a few bridges, as a structural bottleneck that impairs evacuation routes \cite{pregnolato2020assessing}. To quantify these structural differences, we characterize each map with a topological score combining two metrics: the number of exits and the average waypoint distance to the nearest exit (see Table~\ref{tab:maps-difficulties}).

\begin{table}[ht!]
\centering
\small
\begin{tabular}{lccc}
\toprule
\textbf{Map Difficulty} & \textbf{Score} & \textbf{Exits} & \textbf{Exit Distance} \\
\midrule
Easy   & $0.30 \pm 0.02$ & $3$ & $3.22 \pm 0.40$ \\
Medium & $0.67 \pm 0.01$ & $1$ & $4.15 \pm 0.30$ \\
Hard   & $0.77 \pm 0.02$ & $1$ & $6.47 \pm 0.39$ \\
\bottomrule
\end{tabular}
\caption{Topology metrics per difficulty (mean $\pm$ std).
}
\label{tab:maps-difficulties}
\end{table}

Beyond map difficulty, each civilian receives an individual difficulty score based on the shortest-path distance from its spawn waypoint to the nearest exit. Table~\ref{tab:map-civilian-distribution} reports the resulting distribution: easier maps predominantly yield Easy and Medium starting positions, while Hard maps concentrate Hard ones. The two scores are therefore correlated by construction and should be read as complementary rather than independent views (details on both computations in Appendix~\ref{app:difficulties-grouping}).

\begin{table}[ht]
\centering
\small
\begin{tabular}{lccc}
\toprule
\multirow{2}{*}{\textbf{Map Difficulty}} & \multicolumn{3}{c}{\textbf{Civilian Starting Position Difficulty}} \\
\cmidrule(lr){2-4}
 & Easy & Medium & Hard \\
\midrule
Easy   & 30.9\% & 68.2\% &  0.9\% \\
Medium & 18.2\% & 77.6\% &  4.2\% \\
Hard   &  1.3\% & 33.3\% & 65.3\% \\
\bottomrule
\end{tabular}
\caption{Civilian starting position difficulty (shortest-path to nearest exit: Easy $<4$ steps, Medium $4$–$6$, Hard $\geq 7$) across the map difficulty levels.}
\label{tab:map-civilian-distribution}
\end{table}

\subsection{Agents and Information}
\label{subsec:agents}

\begin{figure}[htb]
    \centering
    \begin{subfigure}{0.32\linewidth}
        \centering
        \includegraphics[width=\linewidth]{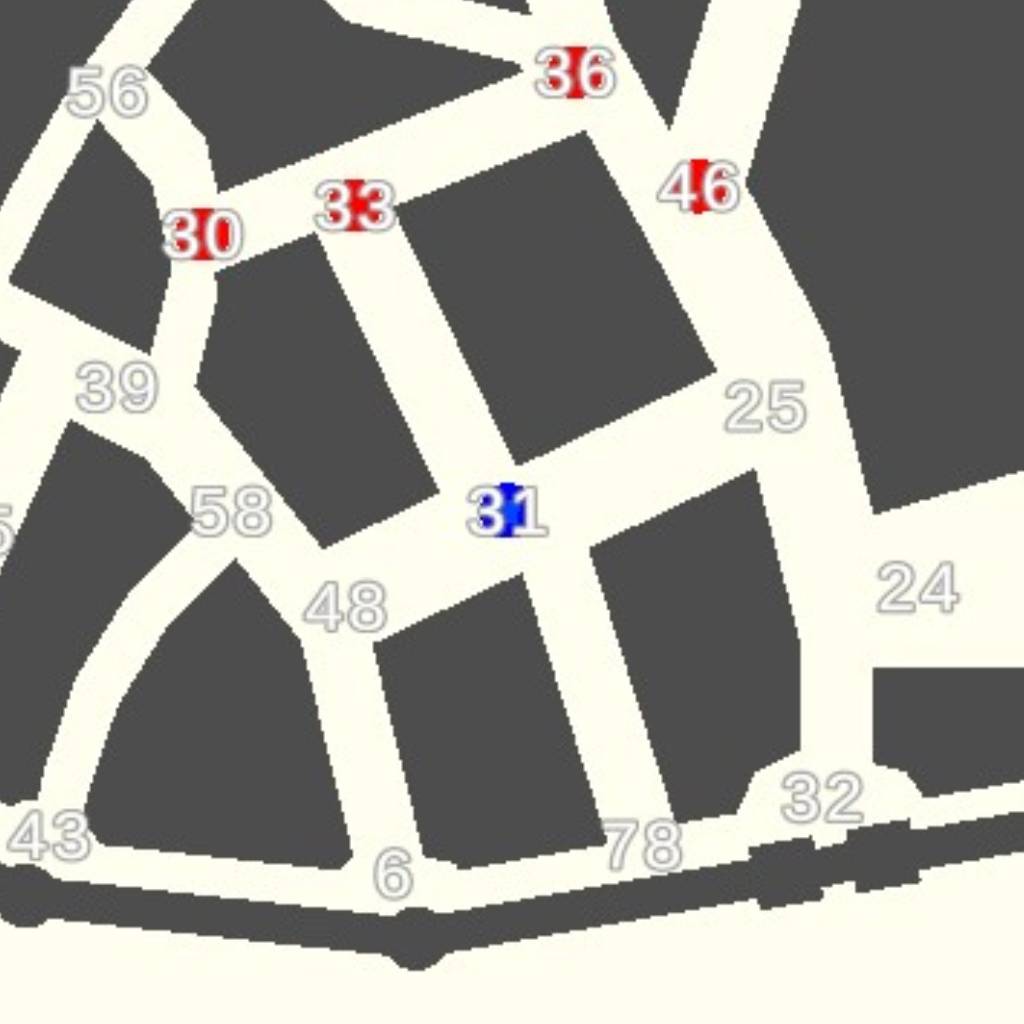}
        \caption{Narrowcast.}
        \label{fig:input1}
    \end{subfigure}
    \hfill
    \begin{subfigure}{0.32\linewidth}
        \centering
        \includegraphics[width=\linewidth]{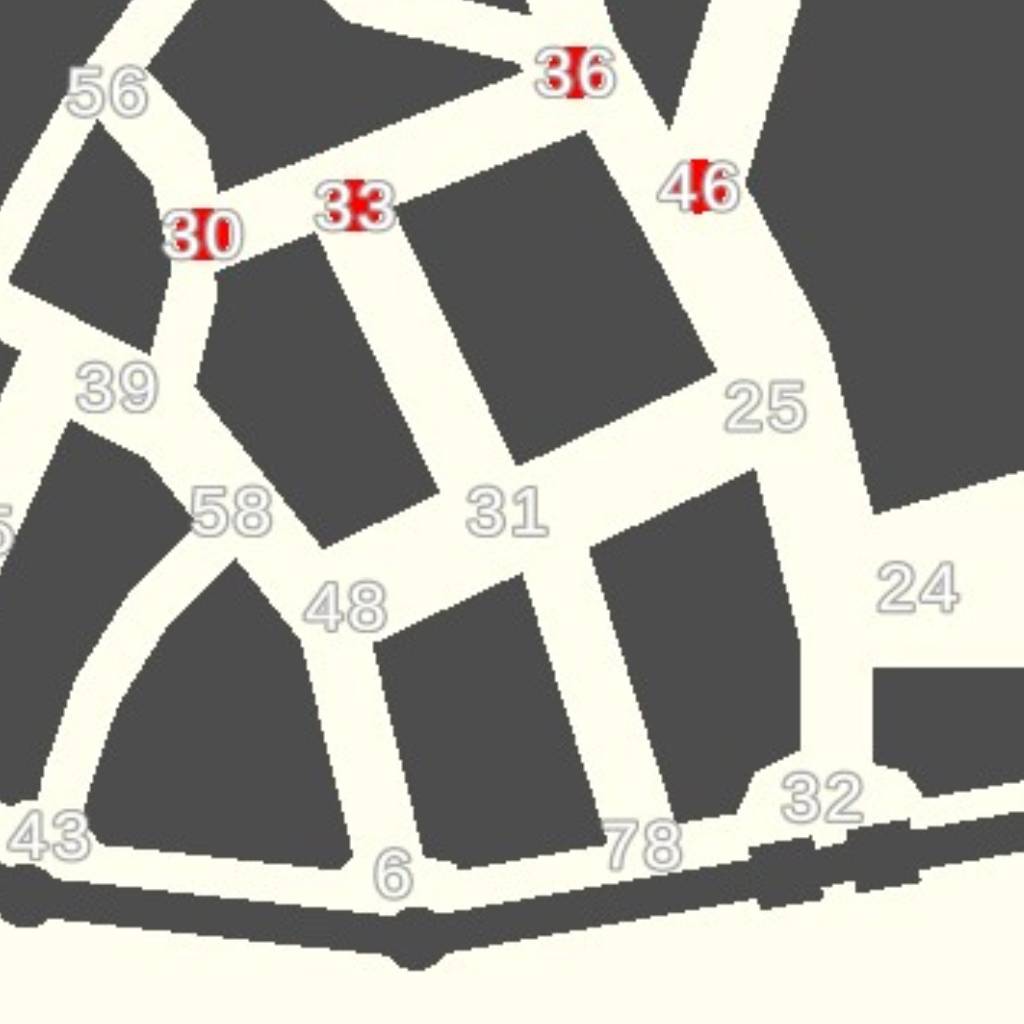}
        \caption{Broadcast.}
        \label{fig:input2}
    \end{subfigure}
    \hfill
    \begin{subfigure}{0.32\linewidth}
        \centering
        \includegraphics[width=\linewidth]{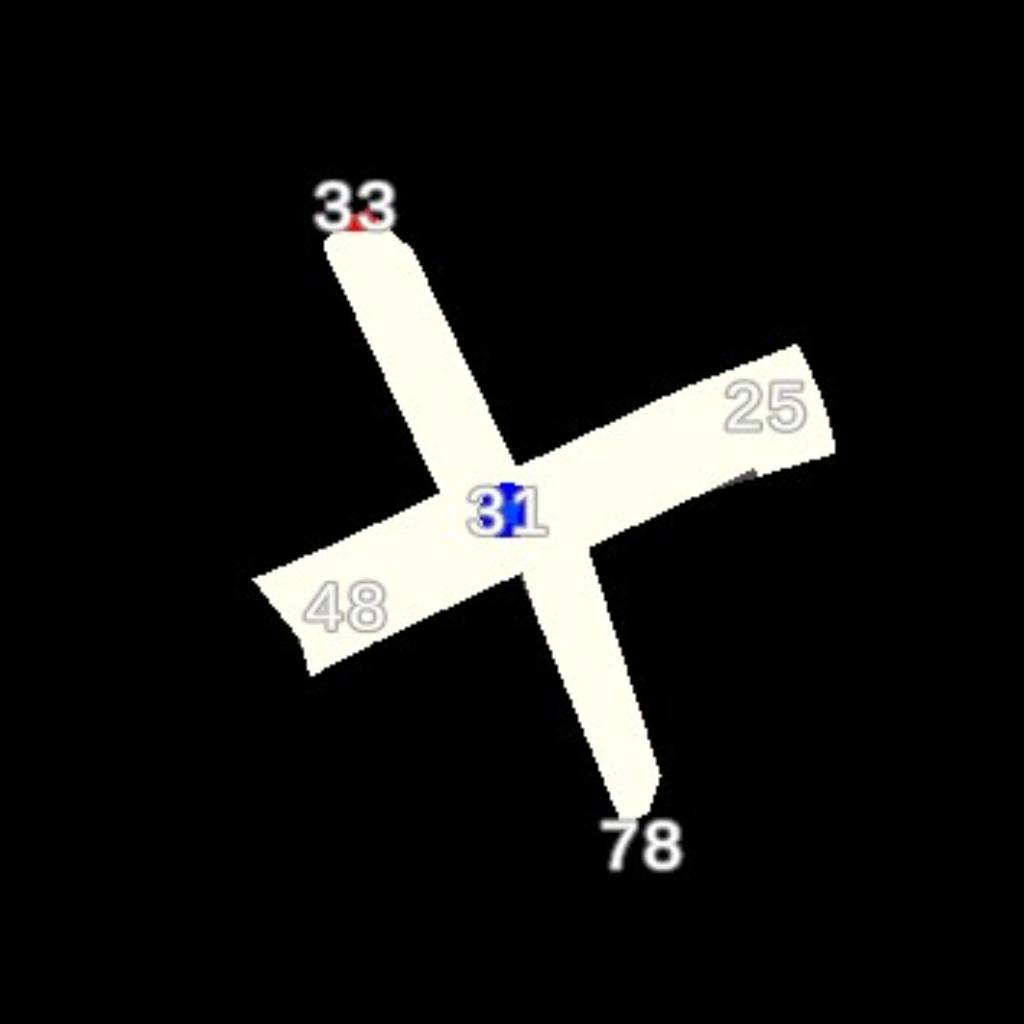}
        \caption{Civilian.}
        \label{fig:input3}
    \end{subfigure}
    \caption{Visual inputs received by each agent type. Operators receive a top-down view of the environment, while civilians have  limited visibility}
    \label{fig:op-civ-input}
\end{figure}

The most natural representation of the environments we just introduced is visual. This requires vision-capable models (VLMs) for the agents that process it. Since by design there is an asymmetry in agents' perception, we now describe the main agent roles (operator and civilian) along with the inputs they receive (see Figure~\ref{fig:op-civ-input}). 

\paragraph{Narrowcast Operator.} The narrowcast operator guides each civilian individually with personalized messages. It receives a top-down view with the target civilian, threats, and waypoints all visible, plus text listing waypoint names for the civilian, threats, and exits, and the waypoints adjacent to the civilian. The operator maintains a separate per-civilian chat history of prior exchanges.

\paragraph{Broadcast Operator.} The broadcast operator issues a single shared message for all civilians, so its inputs are aggregated rather than individualized. The visual input is a top-down view with threats and waypoints visible (civilians are hidden to avoid leaking narrowcast information). The accompanying text lists exit names and threat-occupied waypoints. The chat history is a single sequence of past broadcast messages.

\paragraph{Civilians.} These agents represent the individuals being evacuated; their visual inputs reflect a limited, first-person perspective. Each civilian receives a top-down view with visibility constrained by obstacles and a fixed sight range; threats appear only when visible. The positional text lists the civilian's current waypoint and adjacent reachable waypoints only. Each civilian maintains its own chat history of operator messages and its prior decisions.

Since the environments are organized in connected waypoints, the image can be substituted by an adjacency graph representation that preserves the environment characteristics. The graph is provided as a textual list of waypoints with their adjacent neighbors. With this alternative, the topological structure needed for navigation decisions is preserved while the geometric detail is lost. This alternative, discussed in Section~\ref{sec:experimental-design}, is used to assess the contribution of the visual modality.

Further details on the exact content and format of each input, across both the image and graph configurations, are provided in Appendix~\ref{app:agents-inputs}.

\subsection{Dynamics}

The simulation unfolds as a sequence of discrete turns. Each turn is organized in three phases: \textit{(i)} the operator observes the current state of the environment and produces guidance, \textit{(ii)} civilians receive their message and, combined with their local observation, decide the next waypoint among those adjacent to their current position, and \textit{(iii)} all movements are applied simultaneously, updating the state of the environment. An episode begins at turn 0, with civilians and threats placed at their initial positions, and ends when every civilian has been resolved. A civilian is resolved under one of three outcomes: Save (\hS), reaching an exit; Fail (\hF), coming into contact with a threat; or Timeout (\hT), hitting the maximum-turn limit without reaching an exit or running into a threat. Once resolved, a civilian is removed from the active set and skipped in all subsequent turns. 

The per-turn flow differs between the two communication strategies, reflecting the communication asymmetry introduced in Section~\ref{subsec:agents}. Below we list the main differences in the flow.

\paragraph{Narrowcast.} At the start of each turn the operator is invoked once per active civilian. For each invocation, it receives the per-civilian inputs described in Section~\ref{subsec:agents} (visual input, positional text, and the chat history with that civilian) and produces a single personalized message. The civilian then combines that message with its own local visual and positional input, and selects its next waypoint.

\paragraph{Broadcast.} To simulate communication limited to essential events in order to avoid warning fatigue \cite{sutton2025overalerting}, the operator is invoked only on communication rounds, whose frequency is a parameter of the experimental design (Section~\ref{sec:experimental-design}). On a communication round, the operator receives the aggregated inputs and produces a single shared message, delivered to every active civilian. On turns without a communication round, no new message is issued and civilians act on the most recently received one. In both cases, civilians select a next waypoint at every turn.

\section{Experimental Design}
\label{sec:experimental-design}

The experiments are organized into batches of episodes where each batch corresponds to a different combination of \textit{communication strategy}, \textit{environment representation} and \textit{threats setting}.
Each episode initializes one of the nine maps with specific initial positions for civilians and threats. We sampled 50 starting configurations, with 3 civilians and 12 threats, per map through a semi-automatic procedure to grant that each civilian has at least one path available to an exit (details in Appendix~\ref{app:initial-positions}). This ensures that, in principle, for each episode we have an upper bound of 100\% Save. Each initial position is replayed twice to reduce variability from model generation, obtaining 100 episodes per map per experimental condition. An episode terminates either when all civilians are resolved (Save/Fail)
or when the maximum turn limit is reached. We set this limit to 20 turns, approximately three times the mean shortest-path distance from any waypoint to its nearest exit across all maps (see Table~\ref{tab:maps-difficulties}).

\paragraph{Communication strategy.}
To address RQ1, we evaluate five communication strategies, two for narrowcast and three for broadcast, that vary in content, addressee, and issue frequency.
\\\noindent\textit{\textbf{Narrowcast-Concise}} (\textbf{NC-C}): the operator issues a \textit{concise} personalized message to each civilian, giving the waypoint to reach and a brief motivation. 
\\\noindent\textit{\textbf{Narrowcast-Detailed}} (\textbf{NC-D}), the operator provides a more \textit{detailed} overview of the situation before concluding with a waypoint suggestion. This tests whether providing context before the suggestions improves the quality of the operator decision. 
\\\noindent\textit{\textbf{Broadcast}} (\textbf{BC-1}, \textbf{BC-3}, \textbf{BC-5}): three configurations that issue a single shared message to all civilians, with a new message released every \textit{1}, \textit{3}, or \textit{5} turns respectively. BC-3 and BC-5 limit communication only to some turns to simulate real conditions where broadcast messages are dispatched intermittently, while BC-1 matches the issuing rate of the Narrowcast configurations, setting a control condition with equal message frequency. 

All experiments are conducted with two VLMs acting as both operator and civilian agents: \texttt{Qwen3-VL-30B-FP8} and \texttt{Gemma-3-27B}, hereafter Qwen and Gemma (more details in Appendix \ref{app:experiments-full-details}).

\begin{figure}[t]
\small
\begin{tcolorbox}[title=\textbf{Narrowcast-Concise (NC-C)}, colback=gray!5, colframe=gray!50, boxsep=2pt, left=4pt, right=4pt]
Move to waypoint 23. It's the safest direction away from immediate threats and begins to move you toward an exit.
\end{tcolorbox}
\begin{tcolorbox}[title=\textbf{Narrowcast-Detailed (NC-D)}, colback=gray!5, colframe=gray!50, boxsep=2pt, left=4pt, right=4pt]
(1) There's a threat very close to your location at waypoint 68. Be aware of threats at 8 and 10 as well.
(2) The nearest safe zone is waypoint 28, but it's quite a distance away.
(3) Move to 23. This moves you away from the immediate threat and begins a path toward safety.
\end{tcolorbox}
\begin{tcolorbox}[title=\textbf{Broadcast (BC)}, colback=gray!5, colframe=gray!50, boxsep=2pt, left=4pt, right=4pt]
EMERGENCY BROADCAST: Danger zones: Central area (9, 51, 60, 30), West side (68, 8), East corridor (12, 53, 74). Isolated threat at 46 \& 36. Safe exit is at waypoint 28, in the southeastern area. Avoid exits near threats! Prioritize 28.
\end{tcolorbox}
\caption{Example operator messages for each communication strategy on the same scenario.}
\label{fig:strategy-examples}
\end{figure}

\paragraph{Threat Settings.}
To investigate RQ2 we define two configurations for the threats on the map:
\\\noindent\textit{\textbf{Static}} setting (\hStatic), in which threats remain fixed at their initial positions, and a \noindent\textit{\textbf{Moving}} setting (\hMoving), in which every threat follows a random walk, moving to an adjacent waypoint with probability $0.3$ and remaining in place with probability $0.7$ at each turn, balancing meaningful threat motion with sufficient stability for the operator to plan over multiple turns.

\begin{table*}[ht]
\centering
\footnotesize
\setlength{\tabcolsep}{3pt}
\begin{minipage}{0.49\textwidth}
\centering
\resizebox{\linewidth}{!}{%
\begin{tabular}{l
  >{\columncolor{gray!20}}c >{\columncolor{gray!20}}c >{\columncolor{gray!20}}c
  ccc
  >{\columncolor{gray!20}}c >{\columncolor{gray!20}}c >{\columncolor{gray!20}}c
  }
\toprule
\multirow{2}{*}{\hMapTable}
  & \multicolumn{3}{>{\columncolor{gray!20}}c}{Easy}
  & \multicolumn{3}{c}{Medium}
  & \multicolumn{3}{>{\columncolor{gray!20}}c}{Hard} \\
\cmidrule(lr){2-4} \cmidrule(lr){5-7} \cmidrule(lr){8-10}
& \hF & \hS & \hT & \hF & \hS & \hT & \hF & \hS & \hT \\
\midrule
NC-C$_{\text{G}}$  & 15.7 & \textbf{65.8} & 18.6 & 17.9 & \textbf{51.4} & 30.7 & 23.2 & \underline{31.1} & 45.7 \\
NC-C$_{\text{Q}}$  & 12.0 & 54.0 & 34.0 & \textbf{12.7} & 40.7 & 46.7 & 19.7 & 27.9 & 52.4 \\
\midrule
NC-D$_{\text{G}}$ & \textbf{9.4} & 54.8 & 35.8 & 13.1 & 43.9 & 43.0 & \textbf{11.9} & 25.2 & 62.9 \\
NC-D$_{\text{Q}}$ & 11.8 & 40.6 & 47.7 & 12.8 & 30.4 & 56.8 & 16.4 & 22.2 & 61.3 \\
\midrule
BC-1$_{\text{G}}$ & 47.2 & 35.8 & 17.0 & 31.6 & 23.8 & 44.7 & 42.7 & 22.9 & 34.4 \\
BC-1$_{\text{Q}}$ & 45.4 & 48.1 &  6.4 & 52.9 & 26.4 & 20.7 & 51.3 & 33.8 & 14.9 \\
\midrule
BC-3$_{\text{G}}$ & 46.3 & 33.3 & 20.3 & 37.1 & 20.6 & 42.3 & 44.7 & 21.3 & 34.0 \\
BC-3$_{\text{Q}}$ & 40.8 & 51.1 &  8.1 & 56.3 & 23.9 & 19.8 & 52.1 & \textbf{34.4} & 13.4 \\
\midrule
BC-5$_{\text{G}}$ & 47.6 & 34.1 & 18.3 & 35.0 & 22.3 & 42.7 & 46.8 & 19.8 & 33.4 \\
BC-5$_{\text{Q}}$ & 45.8 & 47.4 &  6.8 & 52.9 & 27.6 & 19.6 & 55.3 & 32.8 & 11.9 \\
\bottomrule
\end{tabular}}
\subcaption{Grouped by map difficulty.}
\label{tab:static-map-difficulty}
\end{minipage}%
\hfill
\begin{minipage}{0.49\textwidth}
\centering
\resizebox{\linewidth}{!}{%
\begin{tabular}{l
  >{\columncolor{gray!20}}c >{\columncolor{gray!20}}c >{\columncolor{gray!20}}c
  ccc
  >{\columncolor{gray!20}}c >{\columncolor{gray!20}}c >{\columncolor{gray!20}}c
  }
\toprule
\multirow{2}{*}{\hCivilianTable}
  & \multicolumn{3}{>{\columncolor{gray!20}}c}{Easy}
  & \multicolumn{3}{c}{Medium}
  & \multicolumn{3}{>{\columncolor{gray!20}}c}{Hard} \\
\cmidrule(lr){2-4} \cmidrule(lr){5-7} \cmidrule(lr){8-10}
& \hF & \hS & \hT & \hF & \hS & \hT & \hF & \hS & \hT \\
\midrule
NC-C$_{\text{G}}$  & 12.6 & \textbf{67.6} & 19.8 & 18.7 & \textbf{53.1} & 28.2 & 24.1 & \underline{27.1} & 48.8 \\
NC-C$_{\text{Q}}$  &  8.8 & 54.2 & 37.0 & 14.3 & 43.3 & 42.4 & 19.9 & 23.3 & 56.8 \\
\midrule
NC-D$_{\text{G}}$ &  \textbf{5.7} & 58.6 & 35.7 & 13.2 & 45.1 & 41.7 & \textbf{11.2} & 19.2 & 69.6 \\
NC-D$_{\text{Q}}$ &  8.8 & 42.3 & 48.9 & \textbf{13.1} & 34.3 & 52.6 & 18.6 & 14.8 & 66.6 \\
\midrule
BC-1$_{\text{G}}$ & 37.9 & 37.4 & 24.7 & 39.2 & 28.0 & 32.8 & 45.4 & 19.1 & 35.5 \\
BC-1$_{\text{Q}}$ & 46.0 & 43.0 & 11.0 & 48.8 & 37.1 & 14.1 & 55.4 & 28.7 & 15.9 \\
\midrule
BC-3$_{\text{G}}$ & 42.9 & 32.6 & 24.5 & 41.7 & 26.2 & 32.1 & 45.1 & 16.7 & 38.2 \\
BC-3$_{\text{Q}}$ & 43.8 & 44.1 & 12.1 & 49.5 & 36.8 & 13.7 & 54.6 & \textbf{30.3} & 15.1 \\
\midrule
BC-5$_{\text{G}}$ & 41.0 & 37.4 & 21.6 & 40.9 & 26.1 & 33.0 & 50.2 & 15.1 & 34.7 \\
BC-5$_{\text{Q}}$ & 47.4 & 41.6 & 11.0 & 49.0 & 37.7 & 13.3 & 60.1 & 27.4 & 12.5 \\
\bottomrule
\end{tabular}}
\subcaption{Grouped by civilian starting position difficulty.}
\label{tab:static-civilian-difficulty}
\end{minipage}
\caption{Outcome rates (\%) for the five communication strategies under static threats. Fail = \hF, Save = \hS, Timeout = \hT. Subscripts denote the operator/civilian model: G = Gemma, Q = Qwen. Results are grouped by (\ref{tab:static-map-difficulty}) map difficulty and (\ref{tab:static-civilian-difficulty}) civilian starting position difficulty.}
\label{tab:static-combined}
\end{table*}

\paragraph{Environment Representation.} 
For RQ3, we further vary the modality of the environment representation provided to the operator, evaluating each modality configuration across all five communication strategies and both models. As introduced in Section~\ref{subsec:agents}, the environment allows both a visual and a graph representation. We compare three operator input configurations: 
\\\noindent\textit{\textbf{Image}}: base setup combining the top-down image with positional text.
\\\noindent\textit{\textbf{Graph}}: image substituted by the graph representation while keeping positional text unchanged.
\\\noindent\textit{\textbf{Image + Graph}}: augments the base setup with the graph representation.

This design allows us to assess whether the graph representation can substitute or complement visual input, providing insight into which modalities are most effective for operator guidance, and whether structured alternatives can compensate for the known limitations of VLMs in visuo-spatial reasoning~\cite{thrush2022winoground}. Civilian inputs are kept fixed across modality conditions, so that any observed variation is attributable to the information available to the operator.

\paragraph{Evaluation.}

Our primary metric is the \textit{outcome rate}: the proportion of civilians resolved as Save, Fail, or Timeout. 
We report them by \textit{map difficulty} (\hMap) and \textit{civilian starting position difficulty} (\hCivilian) to show how they respond to scenario complexity. Save is the desired result, Timeout is an acceptable fallback (since the civilian is unharmed within the simulation span), and Fail is the worst case, representing irreversible harm. We therefore treat Fail as the main performance indicator.

\section{Results}
\label{sec:results}

In this section we report the main results of our experiments for all communication strategies under static threats, moving threats, and input configurations (Table~\ref{tab:static-combined}, Table~\ref{tab:moving-combined}, and Table~\ref{tab:modality-combined} respectively).\footnote{Per-difficulty breakdowns of the modality results are provided in Appendix~\ref{app:modality-full-results}.}

\subsection{Static Threats} 

\paragraph{Fail Rates.}
In this setting, Narrowcast yields the lowest Fail rates: across all difficulties and models, both NC-C and NC-D outperform the Broadcast baselines. The largest NC-C vs BC gaps occur on \textit{Easy} ($+30.6pp$ with Gemma and $+28.8pp$ for Qwen) and remain substantial on \textit{Medium}/\textit{Hard}. NC-D widens the gap further on \textit{Hard} ($+11.3pp$ for Gemma and $+3.3pp$ for Qwen). Grouping by starting position yields the same results but with a steeper Fail-rate as difficulty increases (NC-D$_{\text{Q}}$: $+4.6pp$ by map vs $+9.8pp$ by start).

\begin{table*}[ht]
\centering
\footnotesize
\setlength{\tabcolsep}{3pt}
\begin{minipage}{0.49\textwidth}
\centering
\resizebox{\linewidth}{!}{%
\begin{tabular}{l
  >{\columncolor{gray!20}}c >{\columncolor{gray!20}}c >{\columncolor{gray!20}}c
  ccc
  >{\columncolor{gray!20}}c >{\columncolor{gray!20}}c >{\columncolor{gray!20}}c
  }
\toprule
\multirow{2}{*}{\hMapTable}
  & \multicolumn{3}{>{\columncolor{gray!20}}c}{Easy}
  & \multicolumn{3}{c}{Medium}
  & \multicolumn{3}{>{\columncolor{gray!20}}c}{Hard} \\
\cmidrule(lr){2-4} \cmidrule(lr){5-7} \cmidrule(lr){8-10}
& \hF & \hS & \hT & \hF & \hS & \hT & \hF & \hS & \hT \\
\midrule
NC-C$_{\text{G}}$  & 45.6 & \textbf{43.8} & 10.7 & 53.0 & \textbf{29.2} & 17.8 & 58.7 & 16.6 & 24.8 \\
NC-C$_{\text{Q}}$  & 48.2 & 33.9 & 17.9 & 50.2 & 18.8 & 31.0 & 56.2 & \underline{16.7} & 27.1 \\
\midrule
NC-D$_{\text{G}}$ & \textbf{39.8} & 40.1 & 20.1 & \textbf{43.7} & 24.1 & 32.2 & \textbf{49.3} & 16.0 & 34.7 \\
NC-D$_{\text{Q}}$ & 52.4 & 27.9 & 19.7 & 48.2 & 14.7 & 37.1 & 52.1 & 14.6 & 33.3 \\
\midrule
BC-1$_{\text{G}}$ & 62.1 & 29.2 &  8.7 & 64.1 & 11.9 & 24.0 & 66.6 & 14.0 & 19.4 \\
BC-1$_{\text{Q}}$ & 66.4 & 31.1 &  2.4 & 81.0 & 13.9 &  5.1 & 76.7 & \textbf{19.0} &  4.3 \\
\midrule
BC-3$_{\text{G}}$ & 66.4 & 27.0 &  6.6 & 67.4 & 11.0 & 21.6 & 73.0 & 12.6 & 14.4 \\
BC-3$_{\text{Q}}$ & 66.3 & 31.3 &  2.3 & 81.0 & 14.0 &  5.0 & 78.4 & 18.9 &  2.7 \\
\midrule
BC-5$_{\text{G}}$ & 68.8 & 25.9 &  5.3 & 71.3 & 12.3 & 16.3 & 74.6 & 11.0 & 14.4 \\
BC-5$_{\text{Q}}$ & 67.1 & 31.2 &  1.7 & 79.3 & 15.8 &  4.9 & 78.8 & 17.7 &  3.6 \\
\bottomrule
\end{tabular}}
\subcaption{Grouped by map difficulty.}
\label{tab:moving-map-difficulty}
\end{minipage}%
\hfill
\begin{minipage}{0.49\textwidth}
\centering
\resizebox{\linewidth}{!}{%
\begin{tabular}{l
  >{\columncolor{gray!20}}c >{\columncolor{gray!20}}c >{\columncolor{gray!20}}c
  ccc
  >{\columncolor{gray!20}}c >{\columncolor{gray!20}}c >{\columncolor{gray!20}}c
  }
\toprule
\multirow{2}{*}{\hCivilianTable}
  & \multicolumn{3}{>{\columncolor{gray!20}}c}{Easy}
  & \multicolumn{3}{c}{Medium}
  & \multicolumn{3}{>{\columncolor{gray!20}}c}{Hard} \\
\cmidrule(lr){2-4} \cmidrule(lr){5-7} \cmidrule(lr){8-10}
& \hF & \hS & \hT & \hF & \hS & \hT & \hF & \hS & \hT \\
\midrule
NC-C$_{\text{G}}$  & 41.4 & \textbf{46.0} & 12.6 & 52.5 & \textbf{32.7} & 14.8 & 59.9 & \underline{11.2} & 28.9 \\
NC-C$_{\text{Q}}$  & 41.4 & 36.8 & 21.8 & 51.4 & 24.0 & 24.6 & \textbf{59.1} & \underline{11.2} & 29.7 \\
\midrule
NC-D$_{\text{G}}$ & \textbf{34.4} & 40.7 & 24.9 & \textbf{44.1} & 29.7 & 26.2 & 51.7 &  9.3 & 39.0 \\
NC-D$_{\text{Q}}$ & 46.3 & 29.5 & 24.2 & 51.2 & 19.4 & 29.4 & 53.6 & 10.5 & 35.8 \\
\midrule
BC-1$_{\text{G}}$ & 59.3 & 28.9 & 11.9 & 63.4 & 18.8 & 17.8 & 70.0 &  9.9 & 20.1 \\
BC-1$_{\text{Q}}$ & 65.6 & 31.7 &  2.6 & 74.9 & 21.3 &  3.7 & 80.5 & \textbf{14.0} &  5.5 \\
\midrule
BC-3$_{\text{G}}$ & 63.4 & 24.9 & 11.7 & 67.5 & 18.1 & 14.3 & 76.6 &  7.9 & 15.6 \\
BC-3$_{\text{Q}}$ & 64.7 & 31.1 &  4.2 & 75.0 & 21.9 &  3.1 & 83.3 & 13.4 &  3.3 \\
\midrule
BC-5$_{\text{G}}$ & 63.0 & 28.9 &  8.2 & 71.7 & 16.5 & 11.9 & 77.4 &  7.4 & 15.3 \\
BC-5$_{\text{Q}}$ & 65.0 & 32.2 &  2.9 & 74.8 & 22.1 &  3.0 & 82.9 & 12.6 &  4.6 \\
\bottomrule
\end{tabular}}
\subcaption{Grouped by civilian starting position difficulty.}
\label{tab:moving-civilian-difficulty}
\end{minipage}
\caption{Outcome rates (\%) for the five communication strategies under moving threats. Fail = \hF, Save = \hS, Timeout = \hT. Subscripts denote the operator/civilian model: G = Gemma, Q = Qwen. Results are grouped by (\ref{tab:moving-map-difficulty}) map difficulty and (\ref{tab:moving-civilian-difficulty}) civilian starting position difficulty.}
\label{tab:moving-combined}
\end{table*}

\paragraph{Save Rates.}
Gemma has higher Save rates than Qwen across all difficulties. On \textit{Easy} and \textit{Medium}, NC-C beats the BC strategies (e.g., NC-C$_{\text{G}}$ vs BC-3$_{\text{G}}$ on \textit{Easy}: $+32.5pp$). On \textit{Hard}, NC-C remains Gemma's best strategy, but for Qwen the BC variants outperform NC-C ($+6.5pp$), bringing Qwen's best BC close to Gemma's NC-C. For NC-D, Gemma follows the same pattern, whereas Qwen's NC-D Save rate is below BC even on \textit{Easy} ($-10.5pp$) and the gap widens on \textit{Hard}. The starting-position grouping shows the same pattern with steeper drops across difficulties (NC-C$_{\text{G}}$: $-40.5pp$ vs $-34.7pp$ by map).

\paragraph{Timeout Rates.}
Timeout rates complement and explain the Save rate patterns. For Qwen, Narrowcast leads to high Timeout: NC-C times out more than BC on every map and increases with difficulty (\textit{Hard}: NC-C$_{\text{Q}}$ $52.4\%$ vs BC-3$_{\text{Q}}$ $13.4\%$). For Gemma, NC-C Timeout are similar to BC on \textit{Easy}, lower on \textit{Medium}, and higher on \textit{Hard}. NC-D yields high Timeout for both models on \textit{Hard}, so many non-Save runs end in Timeout rather than failure, producing lower Fail rates. 

\subsection{Moving Threats}

\paragraph{Overall Trends.}

The moving threat setting largely preserves the static threat trends. Fail rates increase in every configuration (e.g., NC-C$_{\text{G}}$ on \textit{Easy}: $15.7\% \to 45.6\%$), but Narrowcast strategies still achieve lower Fail rates than Broadcast (e.g., up to $+20.8pp$ for Gemma on \textit{Easy}). Qwen NC-C keeps its relatively low Fail rates on \textit{Medium} and \textit{Hard}, while Gemma keeps its Save rate advantage ($+10.4pp$ on \textit{Medium}). On \textit{Hard}, Gemma NC-C and Qwen's best BC have similar Save rates but different Fail rates with slightly better Save rate for Qwen's BC, though with a smaller margin ($+2.3pp$ in moving and $+6.5pp$ in static setting). Timeout rates also drop across configurations (e.g., $-20.9pp$ for NC-C$_{\text{G}}$ on \textit{Hard}). The starting-position view confirms the same patterns under moving threats, with a sharper Save-rate drop (NC-C$_{\text{G}}$: $-34.8pp$ vs $-27.2pp$ by map).

\paragraph{Outcome Shifts (Save/Timeout to Fail).}
The increase in Fail rates comes from reductions in both Save and Timeout outcomes. In the static-threat setting, civilians can still avoid fixed threats by wandering, leading to more Timeouts as difficulty increases. With moving threats, these same runs are more likely to end in Fail, converting either Timeouts or Saves to Fail depending on which outcome was more common in the static setting. For NC-C, the higher Fail rate comes mostly from Timeouts on \textit{Hard} maps and from Saves on \textit{Easy} maps, where Timeouts are already low. NC-D shifts more sharply toward Timeouts because static Timeout rates were already high, with Timeouts accounting for most of the increase even on \textit{Easy} for Qwen and dominating on \textit{Hard} for both models.For BC instead, the conversion is Save-driven for Qwen and Timeout-driven for Gemma. The same conversion pattern holds under the starting-position grouping, with comparable Save/Timeout shares.

\subsection{Input Representation}

\begin{table*}[ht]
\centering
\footnotesize
\setlength{\tabcolsep}{3pt}
\begin{minipage}{0.49\textwidth}
\centering
\resizebox{\linewidth}{!}{%
\begin{tabular}{l 
  >{\columncolor{gray!20}}c >{\columncolor{gray!20}}c >{\columncolor{gray!20}}c
  ccc
  >{\columncolor{gray!20}}c >{\columncolor{gray!20}}c >{\columncolor{gray!20}}c
  }
\toprule
\multirow{2}{*}{\hStaticTable}
  & \multicolumn{3}{>{\columncolor{gray!20}}c}{Graph}
  & \multicolumn{3}{c}{Image}
  & \multicolumn{3}{>{\columncolor{gray!20}}c}{Image + Graph} \\
\cmidrule(lr){2-4} \cmidrule(lr){5-7} \cmidrule(lr){8-10}
& \hF & \hS & \hT & \hF & \hS & \hT & \hF & \hS & \hT \\
\midrule
NC-C$_{\text{G}}$  & 36.9 & 23.3 & 39.9 & 18.9 & \textbf{49.4} & 31.6 & 18.6 & 21.7 & 59.7 \\
NC-C$_{\text{Q}}$  & 47.6 & 33.0 & 19.4 & \textbf{14.8} & 40.9 & 44.4 & 45.9 & 38.6 & 15.5 \\
\midrule
NC-D$_{\text{G}}$ & 15.3 & 18.1 & 66.6 & 11.5 & \textbf{41.3} & 47.2 &  \textbf{8.8} & 18.1 & 73.1 \\
NC-D$_{\text{Q}}$ & 31.7 & 37.3 & 31.0 & 13.7 & 31.1 & 55.3 & 24.1 & 43.8 & 32.1 \\
\midrule
BC-1$_{\text{G}}$ & 56.0 & 17.3 & 26.7 & \textbf{40.5} & 27.5 & 32.0 & 43.6 & 25.3 & 31.1 \\
BC-1$_{\text{Q}}$ & 66.7 & 22.2 & 11.1 & 49.9 & 36.1 & 14.0 & 61.5 & 28.7 &  9.9 \\
\midrule
BC-3$_{\text{G}}$ & 55.4 & 16.2 & 28.4 & 42.7 & 25.1 & 32.2 & 44.5 & 23.3 & 32.2 \\
BC-3$_{\text{Q}}$ & 65.8 & 23.0 & 11.3 & 49.7 & \textbf{36.5} & 13.8 & 61.5 & 28.3 & 10.2 \\
\midrule
BC-5$_{\text{G}}$ & 55.1 & 18.3 & 26.6 & 43.1 & 25.4 & 31.5 & 44.0 & 23.0 & 33.0 \\
BC-5$_{\text{Q}}$ & 65.4 & 23.1 & 11.5 & 51.3 & 35.9 & 12.7 & 62.2 & 27.9 & 10.0 \\
\bottomrule
\end{tabular}}
\subcaption{Static threats.}
\label{tab:static-modality}
\end{minipage}%
\hfill
\begin{minipage}{0.49\textwidth}
\centering
\resizebox{\linewidth}{!}{%
\begin{tabular}{l 
  >{\columncolor{gray!20}}c >{\columncolor{gray!20}}c >{\columncolor{gray!20}}c
  ccc
  >{\columncolor{gray!20}}c >{\columncolor{gray!20}}c >{\columncolor{gray!20}}c
  }
\toprule
\multirow{2}{*}{\hMovingTable}
  & \multicolumn{3}{>{\columncolor{gray!20}}c}{Graph}
  & \multicolumn{3}{c}{Image}
  & \multicolumn{3}{>{\columncolor{gray!20}}c}{Image + Graph} \\
\cmidrule(lr){2-4} \cmidrule(lr){5-7} \cmidrule(lr){8-10}
& \hF & \hS & \hT & \hF & \hS & \hT & \hF & \hS & \hT \\
\midrule
NC-C$_{\text{G}}$  & 68.6 & 13.7 & 17.7 & 52.4 & \textbf{29.9} & 17.7 & 63.3 & 13.9 & 22.9 \\
NC-C$_{\text{Q}}$  & 68.4 & 23.4 &  8.2 & \textbf{51.6} & 23.1 & 25.3 & 65.1 & 28.9 &  6.0 \\
\midrule
NC-D$_{\text{G}}$ & 58.7 & 10.1 & 31.2 & \textbf{44.3} & 26.7 & 29.0 & 49.8 & 12.2 & 38.0 \\
NC-D$_{\text{Q}}$ & 61.7 & 24.9 & 13.5 & 50.9 & 19.0 & 30.0 & 55.7 & \textbf{29.7} & 14.6 \\
\midrule
BC-1$_{\text{G}}$ & 77.7 & 10.5 & 11.8 & \textbf{64.3} & 18.4 & 17.4 & 69.1 & 16.3 & 14.6 \\
BC-1$_{\text{Q}}$ & 79.6 & 16.4 &  4.0 & 74.7 & 21.3 &  4.0 & 75.7 & 20.4 &  3.9 \\
\midrule
BC-3$_{\text{G}}$ & 78.1 & 10.3 & 11.5 & 69.0 & 16.9 & 14.2 & 72.1 & 14.7 & 13.2 \\
BC-3$_{\text{Q}}$ & 79.0 & 17.5 &  3.6 & 75.3 & 21.4 &  3.3 & 76.0 & 20.7 &  3.4 \\
\midrule
BC-5$_{\text{G}}$ & 80.4 &  9.5 & 10.1 & 71.6 & 16.4 & 12.0 & 73.6 & 14.6 & 11.8 \\
BC-5$_{\text{Q}}$ & 80.3 & 15.7 &  4.0 & 75.1 & \textbf{21.6} &  3.4 & 77.5 & 19.8 &  2.7 \\
\bottomrule
\end{tabular}}
\subcaption{Moving threats.}
\label{tab:moving-modality}
\end{minipage}
\caption{Outcome rates (\%) across the three operator input configurations, averaged across all maps. \textit{\textbf{Image}} is the baseline setup; \textit{\textbf{Graph}} replaces the image with the graph representation; \textit{\textbf{Image + Graph}} adds the graph on top of the image. Fail = \hF, Save = \hS, Timeout = \hT. Subscripts denote the operator/civilian model: G = Gemma, Q = Qwen. Results are reported under (\ref{tab:static-modality}) static threats and (\ref{tab:moving-modality}) moving threats.}
\label{tab:modality-combined}
\end{table*}

\paragraph{Representation Effects.}
Compared to \textit{\textbf{Graph}}, \textit{\textbf{Image}} improves NC-C performance for both models, reducing Fail rates and increasing Save rates (e.g., Fail $-18.0pp$, Save $+26.1pp$ on Gemma). For Gemma, this recovery translates into Saves, whereas for Qwen the Fail reduction is mostly absorbed by higher Timeouts (Fail $-32.8pp$, Timeout $+25.0pp$). NC-D shows the same pattern: Gemma converts Timeouts into Saves, while Qwen again shifts Fails into Timeouts. In BC, both models exhibit lower Fail and higher Save rates (e.g., Fail $-16.8pp$, Save $+13.9pp$ for BC-1$_{\text{Q}}$). Similar but weaker trends appear in the moving-threat setting.

\paragraph{Model-Specific Effects.}
Adding the graph representation to the image produces opposite effects for Gemma and Qwen. In Narrowcast, Gemma shifts Saves into Timeouts with little change in Fails (e.g.\ NC-C$_{\text{G}}$: Save $-27.7pp$, Timeout $+28.1pp$). Qwen instead reduces Timeouts, reallocating them differently according to the communication strategy: in NC-C$_{\text{Q}}$ mostly into Fails (Timeout $-28.9pp$, Fail $+31.1pp$), while in NC-D$_{\text{Q}}$ mainly into Saves, making \textit{\textbf{Image + Graph}} the best Qwen setup for Save rate. In Broadcast, effects are smaller: Gemma remains comparable to \textit{\textbf{Image}}, while Qwen shows a moderate drop in Saves. Similar trends hold in the moving-threat setting, though part of Gemma’s redistribution shifts from Timeouts to Fails.

\section{Discussion}

\textbf{RQ1} is answered positively. NC strategies consistently achieve lower Fail rates across all configurations and the highest Save rates on \textit{Easy} and \textit{Medium} maps for both models. On \textit{Hard} maps, Gemma's NC-C achieves Save rates comparable to the best BC configuration for Qwen, while maintaining a lower Fail rate, making it preferable from a safety perspective. NC-D pushes this trade-off further by reducing Fail rates even more than NC-C, mainly in favor of higher Timeout rates.\footnote{Manual inspection traced part of these Timeouts to looping behavior, see Appendix~\ref{app:looping} for full details.}

To assess whether these trends extend beyond open-weight operators, we additionally evaluated GPT-5.4 (full details in Appendix~\ref{app:gpt-results}). The Narrowcast advantage is confirmed and improves: Fail rates stay lower than Broadcast and Save rates are higher. This holds also on \textit{Hard} maps, where Narrowcast Save rates now exceed Broadcast (NC-D: 65.6\% vs.\ BC: 54.6\%) instead of just matching them as in open-weight settings. With this stronger model, NC-D also surpasses NC-C, becoming the best strategy overall (Save: 65.6\% vs.\ 57.7\%).

Turning to \textbf{RQ2}, Narrowcast maintains its advantage over Broadcast even when threats move, showing that individualized guidance is robust to threat dynamics. This setting is more challenging for the VLM operator for two reasons: (i) civilians may now enter waypoints that become reachable by a threat within the same turn, and this is not consistently predicted. Additionally, (ii) the longer civilians roam in the map, the more likely they are to encounter a moving threat (most pronounced on static configurations that are prone to Timeouts). 

Experiments addressing \textbf{RQ3} reveal an asymmetric response across the two models. Visual information remains essential and constitutes the most effective environment representation for this task; however, Gemma and Qwen respond differently to the addition of structured textual information with graph representations. For Gemma, the graph representation increases looping behaviour in Narrowcast configurations, shifting outcomes from Saves to Timeouts. In contrast, for Qwen, the graph representation suppresses Timeouts, increasing Fails in NC-C while improving Saves in NC-D. Consequently, NC-D$_{\text{Q}}$ with \textit{\textbf{Image + Graph}} is the only configuration in which the graph representation provides a clear benefit. In all other cases, the graph either has small impact or degrades performance, indicating that visual information cannot be fully replaced by graph-based representations. 

\section{Conclusion}

We present a benchmarking framework for evaluating Vision-Language Model (VLM) operators in crisis evacuation scenarios, where operators guide civilian VLM agents through nine maps of increasing structural complexity toward an exit. Within this framework, we investigated whether individualized guidance performs better than shared communications (Narrowcast vs.\ Broadcast), how environment representation shapes operator behavior (visual vs.\ graph-based), and how threat dynamics affect both (static vs.\ moving).

Based on our findings, Narrowcast consistently reduces Fail rates over Broadcast, an advantage that persists under moving threats. Visual input is essential, while graph representations help only in specific model–strategy combinations and often hurt. Results show that effective guidance depends jointly on strategy and input design, with Narrowcast and visual input being the strongest combination for VLM operators in crisis evacuation.

\section*{Limitations}
Our framework is intended as a first step toward benchmarking VLM-guided evacuation, laying a controlled and reproducible foundation for future work; as a result, it makes several design choices that also bound the scope of our conclusions. Civilians are modeled as homogeneous agents differing only in position and local view, so that performance differences can be attributed to the operator's strategy rather than to civilian heterogeneity. For the same reason, communication is unidirectional: adding dialogue would entangle guidance quality with the ability to repair misunderstandings through follow-up. Our results should therefore be read as an evaluation of operator behavior under controlled conditions, not as a prediction of real-world response distributions.
The operator's top-down view is a best-case observability assumption that measures the upper bound of VLM performance before the degradation introduced by partial sources such as camera feeds; still the task remains challenging even in this setting. The simulation also advances in discrete synchronous turns, abstracting away timing effects such as message latency and overlapping movements. Finally, threat behavior is restricted to static placement and random walks, and does not capture dynamics such as pursuit or danger zone propagation. Relaxing these design choices (heterogeneous civilians, bidirectional dialogue, partial observability, asynchronous timing, and richer threat behavior) is a natural next step toward closing the gap with real evacuation settings.

\section*{Ethical Considerations}

Our benchmarking framework is meant for research purposes only, and does not constitute a field-ready deployment system. Given the high stakes of crisis response and emergency evacuation, the observed civilian failure rates underscore that current Vision-Language Models (VLMs) are not yet reliable enough to operate autonomously in real-world scenarios. Instead we believe that, following a human-AI collaboration paradigm, this technology will be best used in a foreseeable future as a collaborative companion or decision-support tool. In fact, a VLM operator can significantly alleviate cognitive load by drafting initial guidance messages or synthesizing spatial environment data, while ensuring that the final verification and critical routing decisions remain under human oversight.
We thus release our codebase and simulated environments strictly to support reproducibility and future benchmarking. 

Finally, we used AI assistants for coding support and language polishing. All research ideas, experimental design, and analysis are our own.

\bibliography{custom}

\clearpage

\appendix

\section{Implementation and Environment Details}
\label{app:implementation}

\begin{figure*}[ht]
    \centering
    \begin{subfigure}[t]{0.31\textwidth}
        \centering
        \includegraphics[width=\linewidth]{images/maps/easy_1_final.pdf}
        \caption{Easy 1.}
    \end{subfigure}
    \hfill
    \begin{subfigure}[t]{0.31\textwidth}
        \centering
        \includegraphics[width=\linewidth]{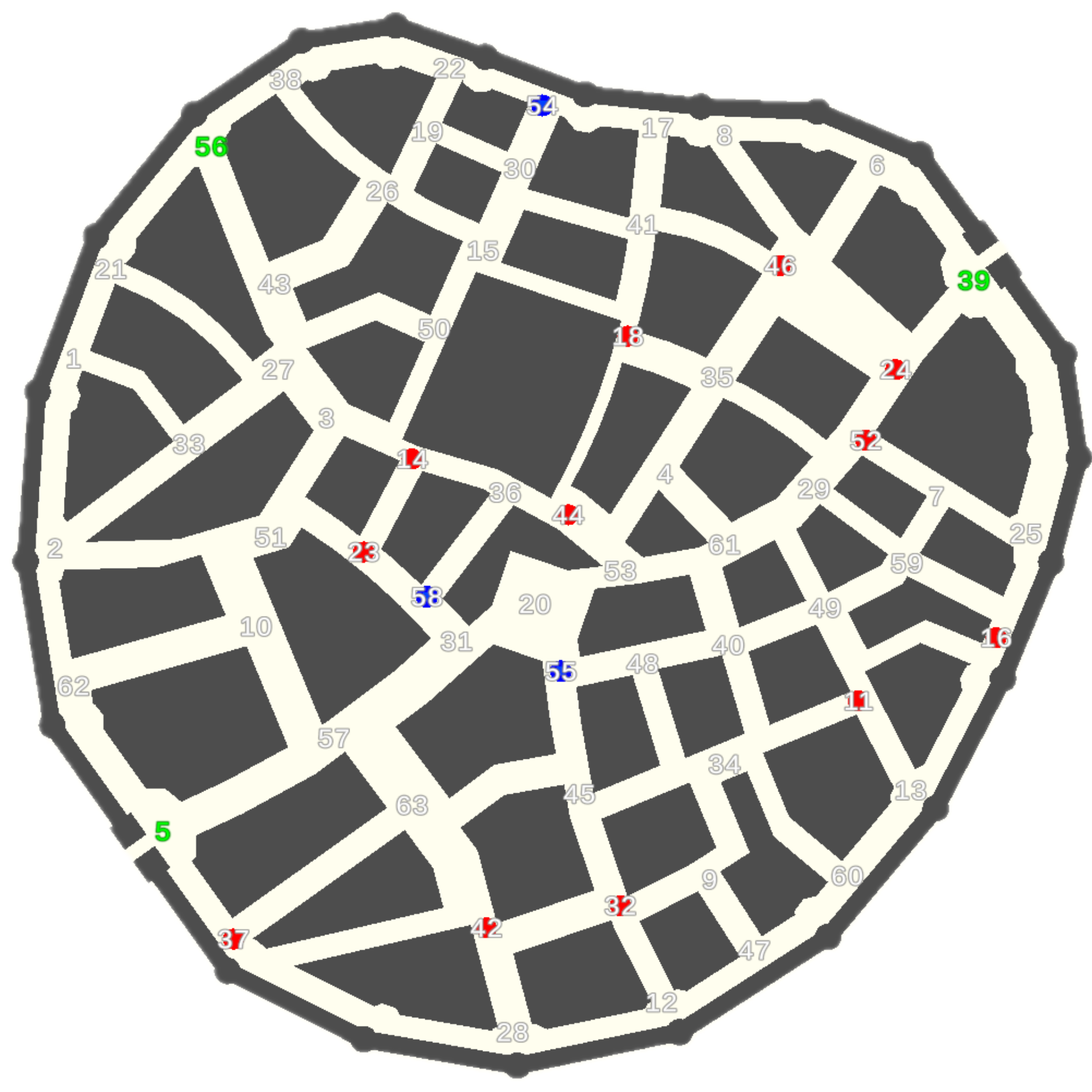}
        \caption{Easy 2.}
    \end{subfigure}
    \hfill
    \begin{subfigure}[t]{0.31\textwidth}
        \centering
        \includegraphics[width=\linewidth]{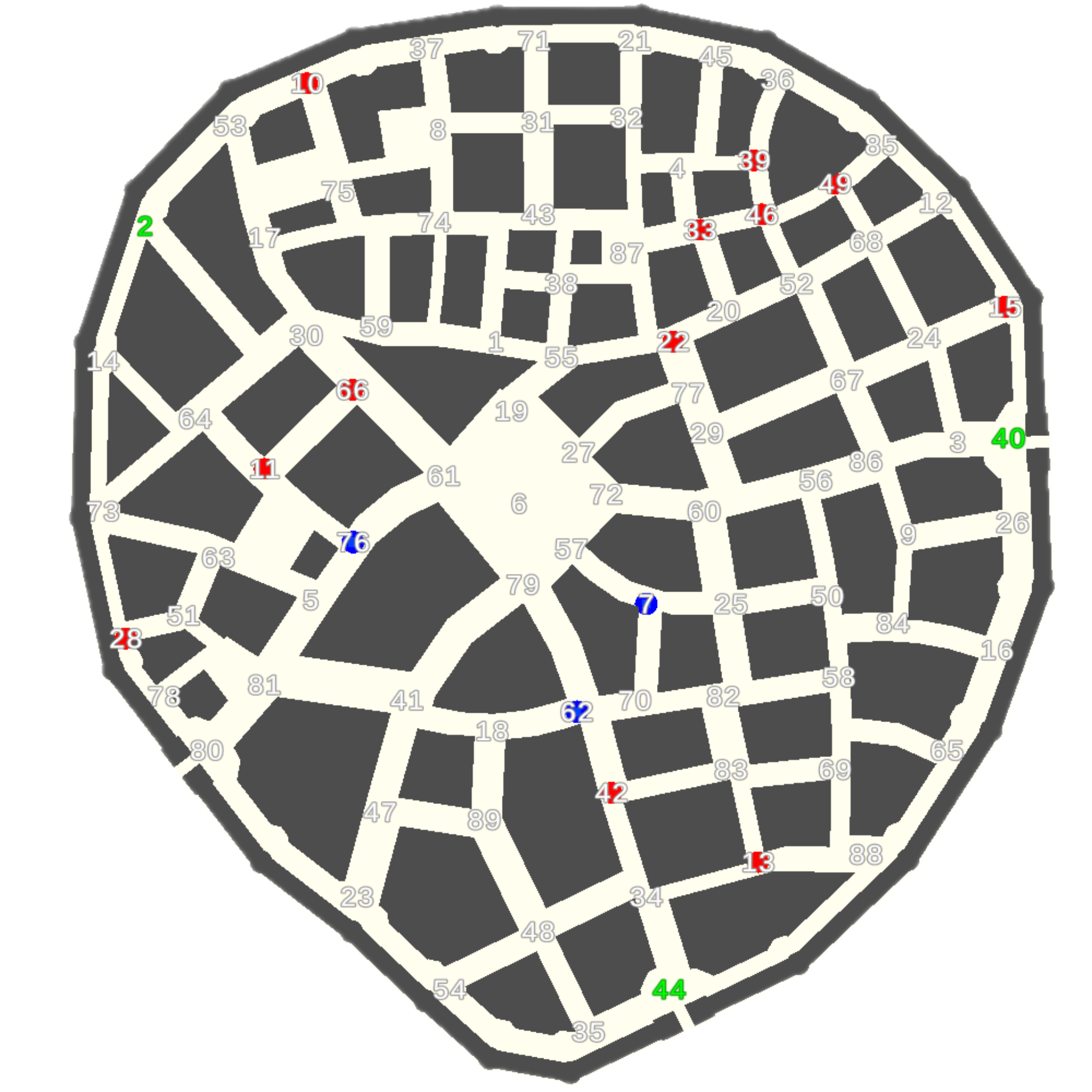}
        \caption{Easy 3.}
    \end{subfigure}
    \\[0.8em]
    \begin{subfigure}[t]{0.31\textwidth}
        \centering
        \includegraphics[width=\linewidth]{images/maps/medium_1_final.pdf}
        \caption{Medium 1.}
    \end{subfigure}
    \hfill
    \begin{subfigure}[t]{0.31\textwidth}
        \centering
        \includegraphics[width=\linewidth]{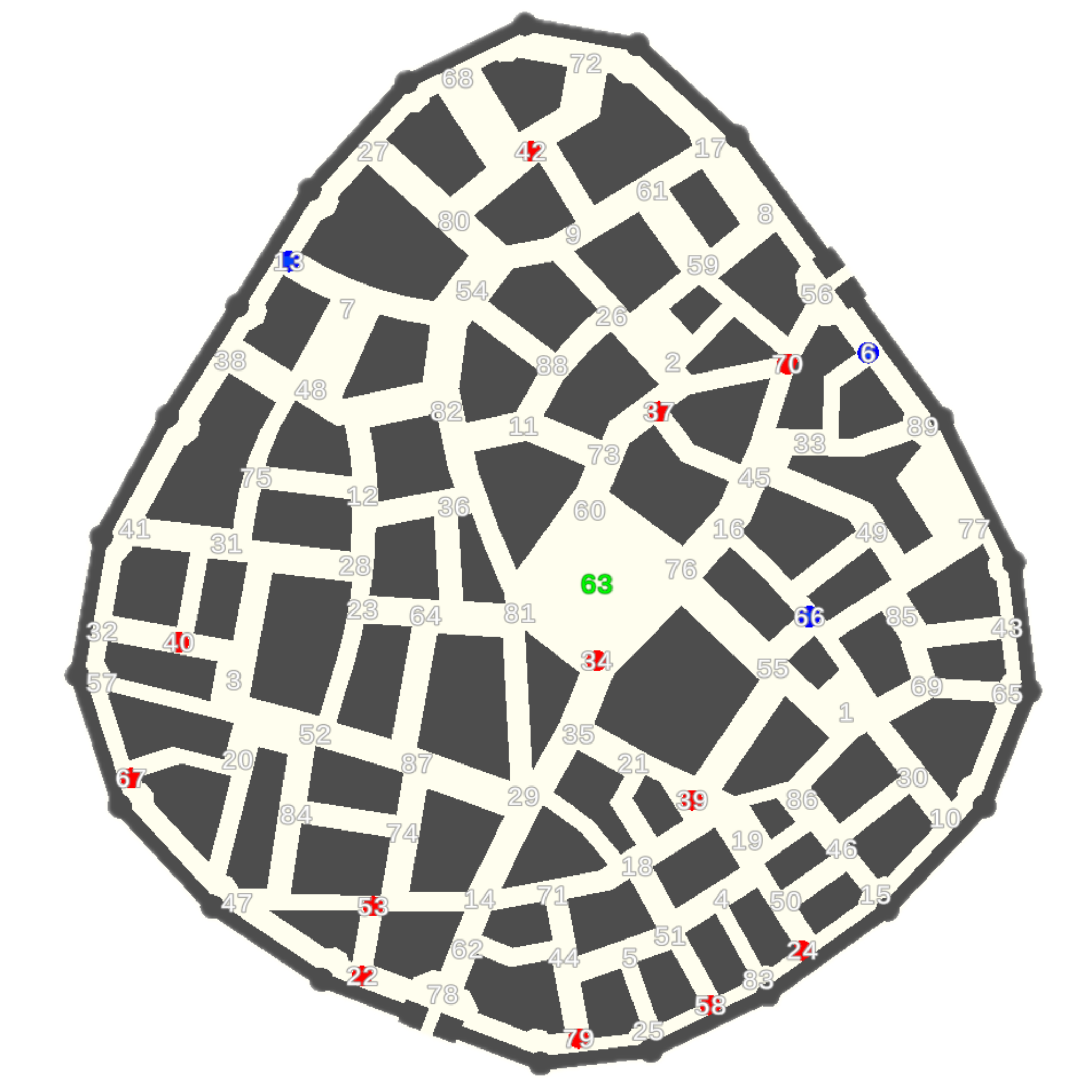}
        \caption{Medium 2.}
    \end{subfigure}
    \hfill
    \begin{subfigure}[t]{0.31\textwidth}
        \centering
        \includegraphics[width=\linewidth]{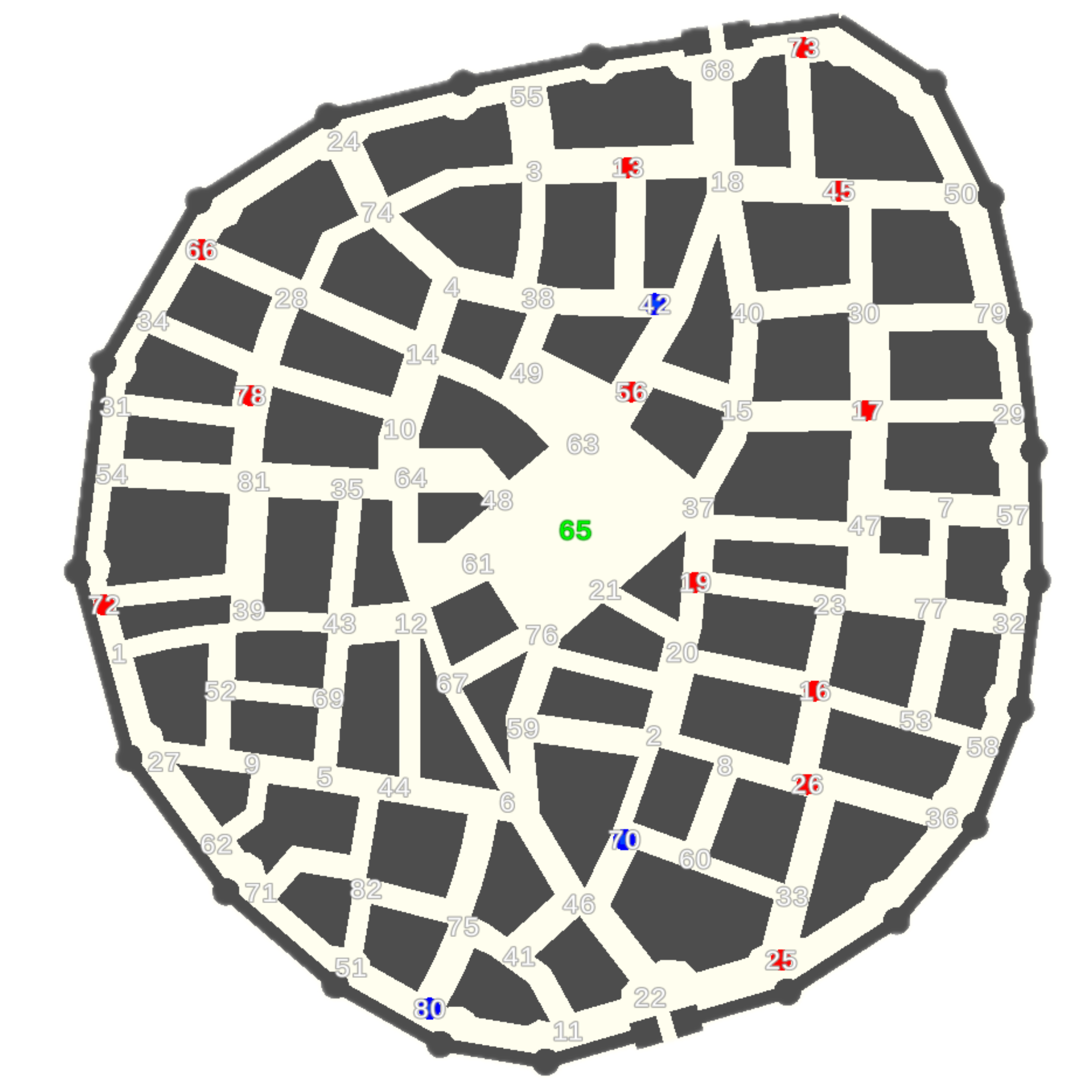}
        \caption{Medium 3.}
    \end{subfigure}
    \\[0.8em]
    \begin{subfigure}[t]{0.31\textwidth}
        \centering
        \includegraphics[width=\linewidth]{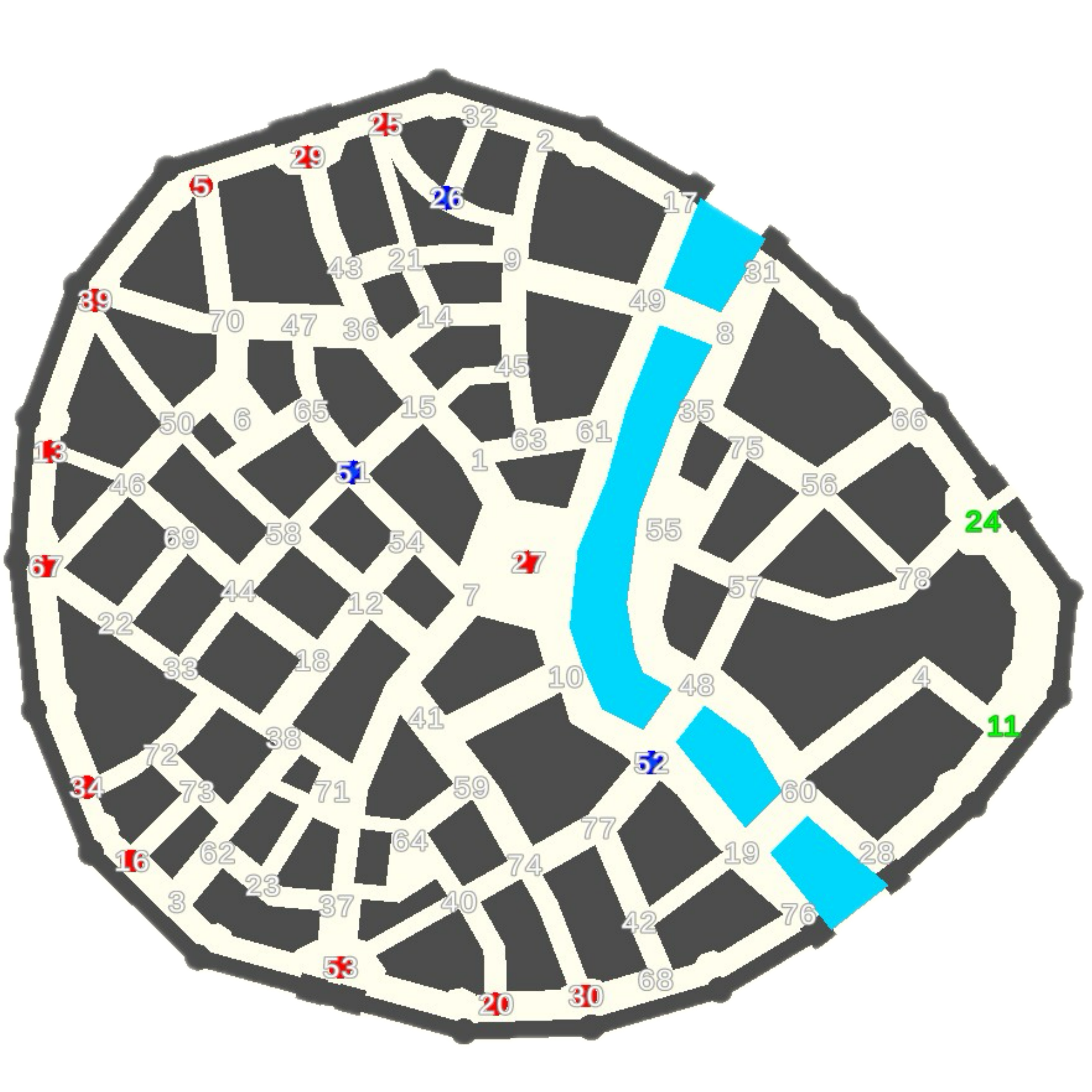}
        \caption{Hard 1.}
    \end{subfigure}
    \hfill
    \begin{subfigure}[t]{0.31\textwidth}
        \centering
        \includegraphics[width=\linewidth]{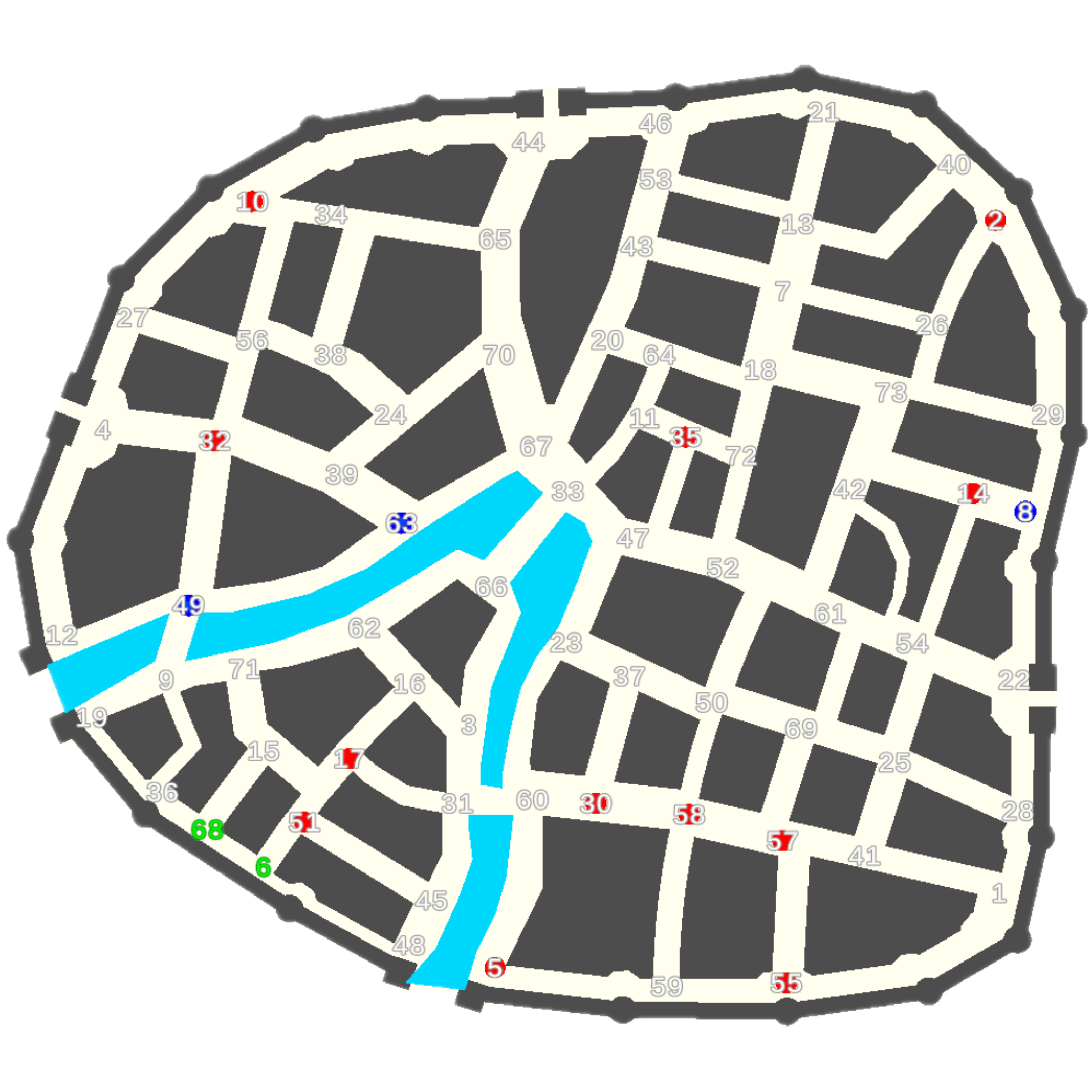}
        \caption{Hard 2.}
    \end{subfigure}
    \hfill
    \begin{subfigure}[t]{0.31\textwidth}
        \centering
        \includegraphics[width=\linewidth]{images/maps/hard_3_final.pdf}
        \caption{Hard 3.}
    \end{subfigure}
\caption{The nine maps used in our experiments. Each row corresponds to a difficulty level: Easy (top), Medium (center), and Hard (bottom). Blue blocks represent civilians, red blocks threats, and green blocks safe exits.}
  \label{fig:all-maps}
\end{figure*}

To create the maps, we started from 2D layouts generated by a procedural town generator,\footnote{\url{https://github.com/watabou/TownGeneratorOS}} exported as \texttt{.svg} files. We then imported these into Blender to extrude buildings into 3D geometry, and finally loaded the resulting models into Unity, where the simulation is run. Figure~\ref{fig:all-maps} shows the full set of nine maps used in our experiments, three per difficulty tier. Topological statistics are reported in Table~\ref{tab:maps-difficulties} of the main text.

In each map, waypoints are labeled with numeric IDs rather than real street or square names for two reasons: (i) real names would not fit within the on-map labels, and (ii) numeric IDs avoid biases stemming from VLM pre-training associations with real-world urban geography. To prevent operators from exploiting label order as an implicit proximity cue, we additionally shuffle the numeric IDs so that consecutive integers do not correspond to adjacent waypoints (see Table~\ref{tab:numeric-bias}). In a real-world deployment, the numeric IDs could be straightforwardly mapped to actual street or square names visible to the operator, without affecting the rest of the system.

\begin{table}[ht]
\centering
\begin{tabular}{lcc}
\toprule
 & \multicolumn{2}{c}{SeqDir\%} \\
\cmidrule(lr){2-3}
Map & Ordered & Shuffled \\
\midrule
Easy 1   & \textbf{62.7\%} & 4.6\% \\
Medium 1 & \textbf{73.5\%} & 7.3\% \\
Hard 1   & \textbf{67.9\%} & 3.7\% \\
\bottomrule
\end{tabular}
\caption{Proportion of turns in which the operator (Qwen) directs civilians along a numeric sequence of waypoints, under ordered versus shuffled labeling.}
\label{tab:numeric-bias}
\end{table}

\section{Difficulty Grouping}
\label{app:difficulties-grouping}

\paragraph{Topological Score.} We characterize each map's structural difficulty using two topology-derived quantities: the average distance civilians must travel to reach an exit, and how many independent exit
clusters the map provides.

\begin{equation}
\text{score} = \frac{1}{2}\left(\frac{\bar{d}(V \setminus E,\, E)}{D_{\max}} + \frac{1}{E_c}\right)                     \end{equation}

\begin{itemize}
\item $\bar{d}(V \setminus E, E)$: mean BFS hop count from non-exit waypoints to the nearest exit.

\item $D_{\max} = 12$: maximum graph diameter observed across all maps (used for normalization).

\item $E_c$: number of connected components in the subgraph induced by exit nodes. Adjacent exits share the same approach path and are counted as one cluster.
\end{itemize}

The first term captures path length difficulty; the second penalizes maps with few distinct exit options. Both terms lie in $[0, 1]$, so the composite score does as well.

\begin{table}[h]
\centering
\begin{tabular}{lccc}
\toprule
\textbf{Map} & \textbf{Exits} & \textbf{Dist to exit} & \textbf{Score} \\
\midrule
Easy-1 & 3 & 3.56 & 0.315 \\
Easy-2 & 3 & 2.78 & 0.283 \\
Easy-3 & 3 & 3.30 & 0.304 \\
\midrule
Medium-1 & 1 & 4.10 & 0.671 \\
Medium-2 & 1 & 4.48 & 0.687 \\
Medium-3 & 1 & 3.88 & 0.662 \\
\midrule
Hard-1 & 1 & 6.47 & 0.770 \\
Hard-2 & 1 & 6.07 & 0.753 \\
Hard-3 & 1 & 6.86 & 0.786 \\
\bottomrule
\end{tabular}
\caption{Per-map topology metrics and composite difficulty score.}
\label{tab:map_scores}
\end{table}

Easy maps have three well-separated exit clusters and short average paths (${\sim}3$ waypoints). Medium maps collapse to a single exit cluster. the exit term alone jumps to $1.0$, with moderate path lengths (${\sim}4$ hops). Hard maps share that single-cluster constraint but add substantially longer paths (${\sim}6$--$7$ hops), compounding the routing challenge. Within-tier score variance is low ($\text{std} \approx 0.01$--$0.02$) relative to the between-tier gaps, confirming that the three tiers are structurally well separated. 

\paragraph{Civilian Spawn Difficulty.} Each civilian is assigned a difficulty based on the Breadth-First Search (BFS) waypoints count from its spawn waypoint to the nearest exit, computed on the topology-only graph (threat positions are not factored in, so the metric is comparable across static and moving-threat conditions).

\begin{equation}
\text{difficulty}(c) =
\begin{cases}
\text{easy}   & d(s_c, E) < 6 \\
\text{medium} & 6 \le d(s_c, E) < 9 \\
\text{hard}   & d(s_c, E) \ge 9
\end{cases}
\end{equation}

\begin{itemize}
\item $s_c$: spawn waypoint of civilian $c$.
\item $d(s_c, E)$: BFS hop count from $s_c$ to the nearest exit waypoint $e \in E$.
\end{itemize}

\section{Agent's Inputs}
\label{app:agents-inputs}

This appendix documents the prompts and per-turn inputs given to each agent in our simulation. We first list the runtime placeholders used to construct prompts (\ref{subsec:placeholders}) and provide representative samples for the two largest of them (\ref{subsec:sample-inputs}). We then describe how multi-turn context is maintained across simulation steps (\ref{subsec:context-handling}), before reproducing the prompt templates for each operator condition (\ref{subsec:nc-c-agent}, \ref{subsec:nc-d-agent}, \ref{subsec:bc-agent}) and for the civilian agent (\ref{subsec:civilian-agent}).

\subsection{Placeholders}
\label{subsec:placeholders}

The prompts below use the following placeholders, substituted at runtime:\\ 
\\
\texttt{\{currentWaypoint\}}: Civilian's current waypoint ID.\\
\texttt{\{adjacentWaypoints\}}: Waypoints reachable in one step from the current position.\\
\texttt{\{safepoints\}}: List of safe exit waypoints.\\
\texttt{\{threatPositions\}}: List of current threat waypoints.\\
\texttt{\{graphText\}}: Textual encoding of the waypoint graph. Populated in \textit{\textbf{Graph}} and \textit{\textbf{Image + Graph}} modalities; empty in \textit{\textbf{Image}}.\\
\texttt{\{operatorMessage\}}: Operator's instruction at the current turn.\\
\texttt{\{fewShotExamples\}}: block of few-shot examples covering both narrowcast and broadcast strategies. The full block contains 4 narrowcast and 3 broadcast examples.

\subsection{Sample inputs}
\label{subsec:sample-inputs}
The following fragments illustrate the runtime content of two placeholders. They are representative excerpts.

\begin{samplebox}[title=\{graphText\}]
Full waypoint graph — all nodes in the map (node: [neighbors]):\\
06 (SAFE EXIT): [19, 31]\\
12: [47, 44, 19]\\
19: [12, 06, 55]\\
31: [47, 06, 55]\\
44: [12, 63, 71]\\
47: [12, 63, 31]\\
63: [47, 44, 71]\\
71: [63, 44]
\end{samplebox}

\begin{samplebox}[title=\{fewShotExamples\}]
NC: Operator: `Move to 2\_2 toward safe exit 0\_3' → \{"command":"goto","target":"2\_2"\}\\
\\
BC: Operator: `Threats near 7, 8, 12. Recommended exit: 26 bottom edge.' → \{"command":"goto","target":"14"\}
\end{samplebox}

\subsection{Context across turns}
\label{subsec:context-handling}

At every simulation turn, each agent is queried with the standard multi-turn chat format: a fixed system prompt followed by an alternating sequence of user and assistant messages, one pair per turn. In the image-enabled modalities (\textit{\textbf{Image}} and \textit{\textbf{Image + Graph}}), the user message of the current turn is also accompanied by an overhead RGB image of the map.

To keep the context window manageable over long simulations, we apply a uniform pruning rule to the chat history: \textit{only the most recent user message carries the token-heavy content}. Concretely, in every prior turn retained in the history we strip (i) the overhead image, (ii) the \texttt{\{graphText\}} block, and (iii) for the civilian agent, the \texttt{\{fewShotExamples\}} block. The textual portion of each prior user message (state placeholders such as \texttt{\{currentWaypoint\}} and \texttt{\{adjacentWaypoints\}}) and the assistant's response are preserved verbatim.

This rule is lossless with respect to the information the agent needs at the current turn: the pruned content describes the present state of the world (image, graph) or the response format (few-shot block), and the freshly injected version at the latest turn already supplies it. Retaining the pruned content in earlier turns would be redundant and would inflate the context with multiple copies of the same information at every step.

\subsection{NC-C Operator}
\label{subsec:nc-c-agent}

\textsc{NC-C} addresses a single civilian per turn and is required only to name the next adjacent waypoint, with no reasoning preamble.

\begin{promptbox}[title=System]
You are an expert emergency operator guiding a civilian to safety in a crisis scenario.\\
\\
You receive an overhead image of the situation and positional data for the civilian.\\
\\
Visual legend: BLUE=civilian, RED=threats (large red disc), WHITE=waypoints, GREEN=safe exits, Black=obstacles, light=walkable terrain, LIGHT BLUE=water (impassable).\\
\\
Your goal:\\
- Help the civilian reach a safe exit (green waypoint) while avoiding threats\\
- Prioritize safety over speed: avoid threats first, then guide toward safety\\
- Give ONE specific adjacent waypoint to move to next\\
\\
Choose the best adjacent waypoint by:\\
1. Avoiding waypoints near threats\\
2. Moving toward the nearest safe exit\\
3. Weigh safety against progress: if a threat blocks the direct path, detour away from the exit to find a safer route. Only push through a risky path if there is no safer alternative. A longer safe path is better than a short dangerous one\\
4. Do not send the civilian back to the waypoint they just came from unless all other moves are unsafe\\
\\
Your response must:\\
- Name the target waypoint clearly
\end{promptbox}

\begin{promptbox}[title=User]
Analyze the image and the positional data, and guide the civilian to safety.\\
\\
Current position: \{currentWaypoint\}\\
Adjacent waypoints: \{adjacentWaypoints\}\\
Safe waypoints: \{safepoints\}\\
Threat positions: \{threatPositions\}\\
\\
\{graphText\}\\
\\
Respond with a short instruction (under 250 chars).
\end{promptbox}

\subsection{NC-D Operator}
\label{subsec:nc-d-agent}

\textsc{NC-D} extends \textsc{NC-C} by requiring an explicit threat and safe-zone preamble before the waypoint choice. Only the blocks that differ from \textsc{NC-C} are shown below; all other content is inherited.

\begin{promptbox-delta}[title=System (replaces final `` response'' block)]
Your response must:\\
- Warn about the nearest threats (skip distant ones that pose no immediate danger)\\
- Indicate the nearest and safest safe zone(s)\\
- Name the suggested target waypoint clearly
\end{promptbox-delta}

\begin{promptbox-delta}[title=User (replaces final instruction line)]
Structure your response in this order:\\
1. Warn about the nearest threats\\
2. Indicate the nearest safe zone(s)\\
3. End with your suggested waypoint (e.g., `Move to X')\\
\\
Respond with a short instruction (under 250 chars).
\end{promptbox-delta}

\subsection{BC Operator}
\label{subsec:bc-agent}

The broadcast operator addresses all civilians on the map with a single shared message at each turn. The message must describe the danger landscape and the recommended exit(s) in terms civilians can interpret from their own local view.

\begin{promptbox}[title=System]
You are an expert emergency operator coordinating multiple civilians in a crisis scenario.\\
\\
You receive an overhead map image and positional data (threat locations and safe exits).\\
\\
Visual legend: RED=threats (large red disc), WHITE=waypoints, GREEN=safe exits, Black=obstacles, light=walkable terrain, LIGHT BLUE=water (impassable).\\
Note: Civilians are NOT visible in this map — you see only threats and waypoints.\\
\\
IMPORTANT: Civilians have LIMITED VISIBILITY — they can only see waypoints near them. They may NOT see the safe exits or distant threats. You must DESCRIBE where the safe exits are located.\\
\\
THREAT REPORTING:\\
- If there are only a FEW threats (1–4): list each threat individually by waypoint name.\\
- If there are MANY threats (5+): group them into DANGER ZONES by area (e.g., `Danger zone in the central area around waypoints 30–35', `Threats clustered along the eastern corridor near waypoints 50, 51, 53'). Mention individual waypoints only for isolated threats outside clusters.\\
\\
SAFE EXIT RECOMMENDATIONS:\\
- Identify which safe exit(s) are the SAFEST to reach.\\
- If some exits are near threats, WARN civilians to avoid those exits.\\
- Clearly define the recommended exit(s) so civilians prioritize them (can also be all of the exits).
\end{promptbox}

\begin{promptbox}[title=User]
Analyze the image and the positional data, and generate ONE broadcast message.\\
\\
Safe waypoints: \{safepoints\}\\
Threat positions: \{threatPositions\}\\
\\
\{graphText\}\\
\\
Your broadcast MUST include:\\
1. [<=4 threats: list individually | 5+ threats: group into DANGER ZONES, summarize clusters]\\
2. Warn about exits near threats and RECOMMEND the SAFEST exit(s) and their location on the map (e.g., `top-left corner', `along the bottom edge')\\
\\
Respond with a short message (under 250 chars).
\end{promptbox}

\subsection{Civilian}
\label{subsec:civilian-agent}

A single civilian prompt is used in all conditions. The operator's instruction is injected through \texttt{\{operatorMessage\}}, and few-shot examples cover both the narrowcast (specific command) and broadcast (general warning) cases so that the civilian agent learns to handle whichever regime is active.

\begin{promptbox}[title=System]
You are a civilian's decision-making brain in a crisis simulation.\\
\\
You receive guidance from an operator, your view of the surroundings, and your positional data.\\
\\
Visual legend: BLUE=you (civilian), RED=threats (large red disc), WHITE Text=waypoints, GREEN Text=safe exits, Black=out of view or obstacles, light=walkable terrain, LIGHT BLUE=water (impassable).\\
\\
You have LIMITED VISIBILITY and may not see all waypoints or threats directly. Use the operator's message alongside your view to make decisions.\\
\\
Your task:\\
- Use the operator's message to choose your next move\\
- Choose only an adjacent waypoint\\
- Avoid returning to the waypoint you just left unless all other moves are unsafe\\
\\
IMPORTANT: The ONLY valid command is ``goto''. Always use ``goto'' as the command.\\
\\
Output format: Return only JSON, no markdown, no prose.\\
\{"command":"goto","target":"<waypointName>"\}
\end{promptbox}

\begin{promptbox}[title=User]
Your current position: \{currentWaypoint\}\\
Adjacent waypoints you can move to: \{adjacentWaypoints\}\\
\\
Operator message: \{operatorMessage\}\\
\\
\{fewShotExamples\}\\
\\
Respond with ONLY the JSON object, nothing else.
\end{promptbox}

\section{Initial Positions}
\label{app:initial-positions}

\paragraph{Positions Generation.} Initial positions of civilians and threats are constructed through a semi-automatic curation process. For each map, the tool samples candidate positions at random and applies a mandatory solvability filter before presenting a scenario for human review.

Specifically, a scenario is accepted as a candidate only if every civilian can reach at least one safe exit via a path that does not pass through any threat spawn position. This is verified by running a BFS from each civilian's waypoint on the threat-blocked graph: if all civilians reach a safe node, the scenario is solvable; otherwise it is discarded and a new sample is drawn. This guarantee holds at spawn time: threats are placed such that no civilian is structurally trapped at the start of the episode, regardless of subsequent threat movement.

Scenarios that pass the solvability check are then displayed in the editor, with civilians and threats moved to their sampled positions so they can be visually inspected the configuration. The human curator can \emph{accept} the scenario (appending it to the saved set and immediately generating the next candidate) or \emph{reject} it (discarding and sampling again). An optional difficulty filter restricts sampling to easy, medium, or hard scenarios based on a composite score, allowing the curator to target a balanced distribution.

\paragraph{Scenario Difficulty Score.}
Each candidate scenario is scored by a composite difficulty metric that combines
two signals derived from the threat-blocked waypoint graph.

\begin{equation}
\text{score} = w_1 \cdot \hat{d}_{\text{civ}} \;+\; w_2 \cdot \hat{\rho}
\end{equation}

\begin{itemize}
\item $\hat{d}_{\text{civ}} = \dfrac{\bar{d}_{\text{blocked}}(C, E)}
      {\max(D, D_{\min})} \in [0,1]$: mean BFS distance from civilian spawn
      positions to the nearest exit, computed on the graph with threat waypoints
      blocked. $D$ is the map diameter and $D_{\min}=10$ is a floor that
      prevents over-amplification on small maps.

\item $\hat{\rho} = \dfrac{\rho_{\max} - 1}{4} \in [0,1]$: normalized maximum
      detour ratio across civilians, where
      $\rho_{\max} = \max_c \bigl(d_{\text{blocked}}(c,E) \,/\, d(c,E)\bigr)$
      measures how much the longest threat-induced detour stretches the optimal
      path. A ratio of $1$ (no detour) maps to $0$; a ratio of $5\times$ or
      more maps to $1$.
\end{itemize}

Default weights are $w_1 = 0.60$, $w_2 = 0.40$ (both terms normalized to $[0,1]$, so the composite score lies in $[0,1]$). The targeted distribution was that most accepted scenarios fell in the score range $[0.35, 0.60]$, and with a slightly larger share below this band than above. This target was informal and not strictly enforced; it served only to discourage extreme over- or under-representation of any difficulty. A small number of scenarios were additionally adjusted by hand after acceptance, for reasons of scenario diversity and curator preference; such edits modified spawn positions only and respected the solvability constraint.

\section{Looping}
\label{app:looping}

\paragraph{Metric Definition.}
cycle$_2$ is computed per civilian per episode. For each step $i \geq 2$, we check whether the operator's target at step $i$ equals the target issued at step $i-2$, i.e.\ whether the civilian was sent back to the waypoint it occupied two turns earlier (A$\to$B$\to$A). cycle$_2$\% is the percentage of such steps over the total number of eligible steps ($i \geq 2$). Values are then aggregated across operator input modality and threat setting.

\paragraph{Analysis.}
In every configuration, looping concentrates in the Timeout column, and NC-D shows higher rates than NC-C, consistent with its higher overall Timeout rates. The model ranking under Narrowcast depends on the operator's environment representation: in the \textit{\textbf{Image}} setting Gemma loops less than Qwen, but this ranking reverses once the graph is introduced (slightly in \textit{\textbf{Graph}} and more sharply in \textit{\textbf{Image + Graph}}), where Gemma loops substantially more. This is in line with Gemma's degradation under graph-augmented inputs reported in Section~\ref{sec:results}. Broadcast strategies present lower looping rates than Narrowcast in most configurations, and follow the opposite model ranking: Gemma loops more than Qwen in every BC setting.

Under moving threats, looping rates in the Timeout column remain comparable to the static setting, but Fail-column rates rise substantially (e.g., in \textit{\textbf{Image}}: NC-C$_{\text{Q}}$ $1.7 \to 23.3$, NC-D$_{\text{Q}}$ $6.4 \to 45.7$). This helps explain the higher Fail rates observed with moving threats: beyond being a structurally harder task, moving threats convert looping civilians into Fails, since civilians cycling between waypoints can be caught by a moving threat rather than safely reaching the turn limit.

\begin{table*}[ht]
\centering
\footnotesize
\setlength{\tabcolsep}{3pt}
\begin{minipage}{0.49\textwidth}
\centering
\resizebox{\linewidth}{!}{%
\begin{tabular}{l
  >{\columncolor{gray!20}}c >{\columncolor{gray!20}}c >{\columncolor{gray!20}}c
  ccc
  >{\columncolor{gray!20}}c >{\columncolor{gray!20}}c >{\columncolor{gray!20}}c
  }
\toprule
\multirow{2}{*}{\hStaticTable}
  & \multicolumn{3}{>{\columncolor{gray!20}}c}{Graph}
  & \multicolumn{3}{c}{Image}
  & \multicolumn{3}{>{\columncolor{gray!20}}c}{Image + Graph} \\
\cmidrule(lr){2-4} \cmidrule(lr){5-7} \cmidrule(lr){8-10}
& \hF & \hS & \hT & \hF & \hS & \hT & \hF & \hS & \hT \\
\midrule
NC-C$_{\text{G}}$  &  5.5 & 5.0 & 34.5 &  2.8 & 1.6 & 16.6 &  7.0 & 3.5 & 53.6 \\
NC-C$_{\text{Q}}$  &  3.7 & 2.2 & 32.2 &  1.7 & 1.2 & 37.1 &  2.3 & 1.6 & 18.8 \\
\midrule
NC-D$_{\text{G}}$  &  3.9 & 3.7 & 45.4 &  6.0 & 2.9 & 37.5 &  4.3 & 3.7 & 67.7 \\
NC-D$_{\text{Q}}$  &  5.6 & 3.0 & 43.9 &  6.4 & 3.4 & 62.3 &  4.2 & 2.1 & 38.9 \\
\midrule
BC-1$_{\text{G}}$  &  2.8 & 1.8 & 20.3 &  2.6 & 1.8 & 18.0 &  3.2 & 2.0 & 16.5 \\
BC-1$_{\text{Q}}$  &  1.4 & 0.9 &  5.7 &  1.2 & 0.7 &  3.4 &  1.1 & 0.4 &  5.0 \\
\midrule
BC-3$_{\text{G}}$  &  3.4 & 1.8 & 19.1 &  3.5 & 1.9 & 19.8 &  3.4 & 2.4 & 17.2 \\
BC-3$_{\text{Q}}$  &  1.3 & 1.0 &  4.5 &  1.2 & 0.6 &  3.3 &  0.8 & 0.8 &  3.3 \\
\midrule
BC-5$_{\text{G}}$  &  3.2 & 2.1 & 15.6 &  3.3 & 2.3 & 18.6 &  3.3 & 2.6 & 16.4 \\
BC-5$_{\text{Q}}$  &  1.2 & 1.1 &  3.7 &  1.3 & 0.5 &  3.4 &  1.1 & 0.7 &  3.1 \\
\bottomrule
\end{tabular}}
\subcaption{Static threats.}
\label{tab:loop-static}
\end{minipage}%
\hfill
\begin{minipage}{0.49\textwidth}
\centering
\resizebox{\linewidth}{!}{%
\begin{tabular}{l
  >{\columncolor{gray!20}}c >{\columncolor{gray!20}}c >{\columncolor{gray!20}}c
  ccc
  >{\columncolor{gray!20}}c >{\columncolor{gray!20}}c >{\columncolor{gray!20}}c
  }
\toprule
\multirow{2}{*}{\hMovingTable}
  & \multicolumn{3}{>{\columncolor{gray!20}}c}{Graph}
  & \multicolumn{3}{c}{Image}
  & \multicolumn{3}{>{\columncolor{gray!20}}c}{Image + Graph} \\
\cmidrule(lr){2-4} \cmidrule(lr){5-7} \cmidrule(lr){8-10}
& \hF & \hS & \hT & \hF & \hS & \hT & \hF & \hS & \hT \\
\midrule
NC-C$_{\text{G}}$  & 15.0 & 5.1 & 38.1 &  7.1 & 1.5 & 15.5 & 34.2 & 3.7 & 58.4 \\
NC-C$_{\text{Q}}$  &  8.9 & 2.3 & 26.6 & 23.3 & 1.8 & 36.9 &  5.0 & 1.1 & 20.0 \\
\midrule
NC-D$_{\text{G}}$  & 34.1 & 3.7 & 53.3 & 25.1 & 2.9 & 38.5 & 54.4 & 4.5 & 70.9 \\
NC-D$_{\text{Q}}$  & 17.3 & 3.1 & 42.9 & 45.7 & 5.6 & 59.3 & 17.8 & 2.8 & 44.2 \\
\midrule
BC-1$_{\text{G}}$  &  7.7 & 1.7 & 22.8 & 11.2 & 1.5 & 21.2 &  9.7 & 2.2 & 20.1 \\
BC-1$_{\text{Q}}$  &  1.6 & 0.6 &  4.5 &  1.5 & 0.6 & 10.0 &  1.5 & 0.9 &  6.5 \\
\midrule
BC-3$_{\text{G}}$  &  7.2 & 2.8 & 17.7 &  9.5 & 2.0 & 20.1 &  9.9 & 2.0 & 19.7 \\
BC-3$_{\text{Q}}$  &  1.4 & 0.7 &  4.0 &  1.5 & 0.5 &  7.9 &  1.2 & 0.5 &  8.9 \\
\midrule
BC-5$_{\text{G}}$  &  6.7 & 2.5 & 15.5 &  6.6 & 2.1 & 17.1 &  7.9 & 2.2 & 17.3 \\
BC-5$_{\text{Q}}$  &  1.7 & 0.9 &  4.9 &  1.7 & 0.9 &  5.4 &  1.0 & 0.5 &  3.8 \\
\bottomrule
\end{tabular}}
\subcaption{Moving threats.}
\label{tab:loop-moving}
\end{minipage}
\caption{Looping analysis: cycle$_2$\% across the three operator input configurations, aggregated over all nine maps. \textit{\textbf{Image}} is the baseline setup; \textit{\textbf{Graph}} replaces the image with the graph representation; \textit{\textbf{Image + Graph}} adds the graph on top of the image. Values are broken down by outcome (Fail = \hF, Save = \hS, Timeout = \hT). Subscripts denote the operator/civilian model: G = Gemma, Q = Qwen.}
\label{tab:loop-combined}
\end{table*}

\section{Modality Interactions Full Results}
\label{app:modality-full-results}

\paragraph{Graph.} The per-difficulty breakdown confirms the average trends reported in Table~\ref{tab:modality-combined}. The Narrowcast advantage over Broadcast on Fail rates holds at every difficulty level for both models, and outcomes degrade consistently as difficulty increases. Without the image, Gemma's Save rates drop sharply, especially under NC-D on Hard (7.0\%), with most non-Fail outcomes absorbed by Timeouts. This is in line with the increased looping behavior reported in Appendix~\ref{app:looping}. Qwen tolerates the graph-only setting better than Gemma under Narrowcast, achieving higher Saves and lower Timeouts at every difficulty. Under moving threats, Fail rates rise across all configurations, but the relative ranking between strategies and models is preserved.

\paragraph{Image + Graph modality.} Difficulty trends mirror those of the other two modalities, with no tier-specific anomaly introduced by adding the graph on top of the image. The Gemma Save-into-Timeout redistribution discussed in Table~\ref{tab:modality-combined} is visible at every difficulty level under both NC-C and NC-D, confirming it as a systematic effect of the added graph rather than a map-specific artifact. Qwen's NC-D gain from the graph holds across all difficulties, with NC-D$_{\text{Q}}$ remaining its best configuration on Save rate also when broken down by map tier. Broadcast results stay roughly stable across difficulties, indicating that the model-specific divergence concentrates in the Narrowcast conditions. Under moving threats, the same patterns persist, with part of Gemma's redistribution shifting from Timeouts into Fails on Hard maps.

\begin{table*}[ht]
\centering
\footnotesize
\setlength{\tabcolsep}{3pt}
\begin{minipage}{0.49\textwidth}
\centering
\resizebox{\linewidth}{!}{%
\begin{tabular}{l
  >{\columncolor{gray!20}}c >{\columncolor{gray!20}}c >{\columncolor{gray!20}}c
  ccc
  >{\columncolor{gray!20}}c >{\columncolor{gray!20}}c >{\columncolor{gray!20}}c
  }
\toprule
\multirow{2}{*}{\hStaticTable}
  & \multicolumn{3}{>{\columncolor{gray!20}}c}{Easy}
  & \multicolumn{3}{c}{Medium}
  & \multicolumn{3}{>{\columncolor{gray!20}}c}{Hard} \\
\cmidrule(lr){2-4} \cmidrule(lr){5-7} \cmidrule(lr){8-10}
& \hF & \hS & \hT & \hF & \hS & \hT & \hF & \hS & \hT \\
\midrule
NC-C$_{\text{G}}$   & 31.9 & 31.8 & 36.3 & 34.7 & 26.2 & 39.1 & 44.1 & 11.8 & 44.1 \\
NC-C$_{\text{Q}}$   & 36.6 & 48.6 & 14.9 & 53.0 & 36.4 & 10.6 & 53.2 & 14.0 & 32.8 \\
\midrule
NC-D$_{\text{G}}$  & 13.7 & 30.8 & 55.6 & 14.4 & 16.6 & 69.0 & 17.7 &  7.0 & 75.3 \\
NC-D$_{\text{Q}}$  & 25.1 & 56.0 & 18.9 & 34.2 & 36.7 & 29.1 & 35.9 & 19.2 & 44.9 \\
\midrule
BC-1$_{\text{G}}$ & 53.2 & 25.7 & 21.1 & 52.9 & 17.0 & 30.1 & 62.0 &  9.2 & 28.8 \\
BC-1$_{\text{Q}}$ & 64.1 & 29.8 &  6.1 & 62.6 & 23.0 & 14.4 & 73.6 & 13.8 & 12.7 \\
\midrule
BC-3$_{\text{G}}$ & 52.4 & 24.4 & 23.1 & 53.8 & 16.6 & 29.7 & 60.0 &  7.6 & 32.4 \\
BC-3$_{\text{Q}}$ & 63.3 & 28.6 &  8.1 & 61.9 & 24.9 & 13.2 & 72.1 & 15.4 & 12.4 \\
\midrule
BC-5$_{\text{G}}$ & 50.9 & 27.2 & 21.9 & 53.3 & 18.3 & 28.3 & 61.0 &  9.4 & 29.6 \\
BC-5$_{\text{Q}}$ & 62.3 & 32.4 &  5.2 & 62.4 & 21.7 & 15.9 & 71.3 & 15.3 & 13.3 \\
\bottomrule
\end{tabular}}
\subcaption{Static threats.}
\label{tab:graph-static}
\end{minipage}%
\hfill
\begin{minipage}{0.49\textwidth}
\centering
\resizebox{\linewidth}{!}{%
\begin{tabular}{l
  >{\columncolor{gray!20}}c >{\columncolor{gray!20}}c >{\columncolor{gray!20}}c
  ccc
  >{\columncolor{gray!20}}c >{\columncolor{gray!20}}c >{\columncolor{gray!20}}c
  }
\toprule
\multirow{2}{*}{\hMovingTable} 
  & \multicolumn{3}{>{\columncolor{gray!20}}c}{Easy}
  & \multicolumn{3}{c}{Medium}
  & \multicolumn{3}{>{\columncolor{gray!20}}c}{Hard} \\
\cmidrule(lr){2-4} \cmidrule(lr){5-7} \cmidrule(lr){8-10}
& \hF & \hS & \hT & \hF & \hS & \hT & \hF & \hS & \hT \\
\midrule
NC-C$_{\text{G}}$   & 64.2 & 20.1 & 15.7 & 66.1 & 14.2 & 19.7 & 75.4 &  6.8 & 17.8 \\
NC-C$_{\text{Q}}$   & 56.4 & 38.1 &  5.4 & 73.1 & 21.3 &  5.6 & 75.6 & 10.8 & 13.7 \\
\midrule
NC-D$_{\text{G}}$  & 56.6 & 17.9 & 25.6 & 58.4 &  9.3 & 32.2 & 61.0 &  3.1 & 35.9 \\
NC-D$_{\text{Q}}$  & 55.1 & 39.0 &  5.9 & 62.1 & 23.1 & 14.8 & 67.8 & 12.4 & 19.8 \\
\midrule
BC-1$_{\text{G}}$ & 76.0 & 16.2 &  7.8 & 77.2 &  9.1 & 13.7 & 79.8 &  6.2 & 14.0 \\
BC-1$_{\text{Q}}$ & 72.9 & 24.8 &  2.3 & 79.9 & 15.6 &  4.6 & 85.9 &  9.0 &  5.1 \\
\midrule
BC-3$_{\text{G}}$ & 74.7 & 16.4 &  8.9 & 79.0 &  9.3 & 11.7 & 80.8 &  5.2 & 14.0 \\
BC-3$_{\text{Q}}$ & 73.6 & 24.6 &  1.9 & 78.6 & 17.7 &  3.8 & 84.8 & 10.2 &  5.0 \\
\midrule
BC-5$_{\text{G}}$ & 77.2 & 15.0 &  7.8 & 78.6 &  8.2 & 13.2 & 85.4 &  5.2 &  9.3 \\
BC-5$_{\text{Q}}$ & 73.3 & 23.8 &  2.9 & 81.4 & 14.4 &  4.1 & 86.1 &  9.0 &  4.9 \\
\bottomrule
\end{tabular}}
\subcaption{Moving threats.}
\label{tab:graph-moving}
\end{minipage}
\caption{\textit{\textbf{Graph}} modality results, grouped by map difficulty. Fail = \hF, Save = \hS, Timeout = \hT. Subscripts denote the operator/civilian model: G = Gemma, Q = Qwen.}
\label{tab:graph-combined}
\end{table*}

\begin{table*}[ht]
\centering
\footnotesize
\setlength{\tabcolsep}{3pt}
\begin{minipage}{0.49\textwidth}
\centering
\resizebox{\linewidth}{!}{%
\begin{tabular}{l
  >{\columncolor{gray!20}}c >{\columncolor{gray!20}}c >{\columncolor{gray!20}}c
  ccc
  >{\columncolor{gray!20}}c >{\columncolor{gray!20}}c >{\columncolor{gray!20}}c
  }
\toprule
\multirow{2}{*}{\hStaticTable}
  & \multicolumn{3}{>{\columncolor{gray!20}}c}{Easy}
  & \multicolumn{3}{c}{Medium}
  & \multicolumn{3}{>{\columncolor{gray!20}}c}{Hard} \\
\cmidrule(lr){2-4} \cmidrule(lr){5-7} \cmidrule(lr){8-10}
& \hF & \hS & \hT & \hF & \hS & \hT & \hF & \hS & \hT \\
\midrule
NC-C$_{\text{G}}$   & 16.9 & 25.7 & 57.4 & 20.1 & 28.6 & 51.3 & 18.9 & 10.8 & 70.3 \\
NC-C$_{\text{Q}}$   & 41.4 & 50.2 &  8.3 & 44.0 & 45.9 & 10.1 & 52.3 & 19.7 & 28.0 \\
\midrule
NC-D$_{\text{G}}$  &  9.3 & 25.8 & 64.9 &  8.1 & 21.3 & 70.6 &  8.9 &  7.1 & 84.0 \\
NC-D$_{\text{Q}}$  & 22.2 & 55.3 & 22.4 & 22.0 & 49.4 & 28.6 & 28.2 & 26.6 & 45.2 \\
\midrule
BC-1$_{\text{G}}$ & 39.7 & 35.3 & 25.0 & 41.4 & 21.9 & 36.7 & 49.6 & 18.8 & 31.7 \\
BC-1$_{\text{Q}}$ & 54.3 & 41.0 &  4.7 & 65.1 & 19.6 & 15.3 & 65.0 & 25.4 &  9.6 \\
\midrule
BC-3$_{\text{G}}$ & 40.7 & 30.9 & 28.4 & 44.8 & 19.3 & 35.9 & 48.1 & 19.6 & 32.3 \\
BC-3$_{\text{Q}}$ & 54.3 & 41.1 &  4.6 & 63.3 & 19.8 & 16.9 & 66.9 & 24.0 &  9.1 \\
\midrule
BC-5$_{\text{G}}$ & 40.1 & 31.4 & 28.4 & 43.1 & 19.9 & 37.0 & 48.9 & 17.4 & 33.7 \\
BC-5$_{\text{Q}}$ & 57.1 & 38.0 &  4.9 & 63.4 & 20.6 & 16.0 & 66.0 & 25.0 &  9.0 \\
\bottomrule
\end{tabular}}
\subcaption{Static threats.}
\label{tab:imggraph-static}
\end{minipage}%
\hfill
\begin{minipage}{0.49\textwidth}
\centering
\resizebox{\linewidth}{!}{%
\begin{tabular}{l
  >{\columncolor{gray!20}}c >{\columncolor{gray!20}}c >{\columncolor{gray!20}}c
  ccc
  >{\columncolor{gray!20}}c >{\columncolor{gray!20}}c >{\columncolor{gray!20}}c
  }
\toprule
\multirow{2}{*}{\hMovingTable}
  & \multicolumn{3}{>{\columncolor{gray!20}}c}{Easy}
  & \multicolumn{3}{c}{Medium}
  & \multicolumn{3}{>{\columncolor{gray!20}}c}{Hard} \\
\cmidrule(lr){2-4} \cmidrule(lr){5-7} \cmidrule(lr){8-10}
& \hF & \hS & \hT & \hF & \hS & \hT & \hF & \hS & \hT \\
\midrule
NC-C$_{\text{G}}$   & 61.7 & 18.3 & 20.0 & 59.9 & 16.2 & 23.9 & 68.3 &  7.0 & 24.7 \\
NC-C$_{\text{Q}}$   & 56.7 & 40.8 &  2.6 & 65.0 & 31.0 &  4.0 & 73.8 & 14.8 & 11.4 \\
\midrule
NC-D$_{\text{G}}$  & 50.4 & 18.7 & 30.9 & 46.1 & 13.2 & 40.7 & 52.9 &  4.7 & 42.4 \\
NC-D$_{\text{Q}}$  & 48.2 & 42.7 &  9.1 & 55.4 & 27.9 & 16.7 & 63.4 & 18.4 & 18.1 \\
\midrule
BC-1$_{\text{G}}$ & 65.0 & 26.1 &  8.9 & 68.9 & 11.9 & 19.2 & 73.4 & 10.9 & 15.7 \\
BC-1$_{\text{Q}}$ & 64.9 & 33.0 &  2.1 & 81.0 & 13.4 &  5.6 & 81.2 & 14.9 &  3.9 \\
\midrule
BC-3$_{\text{G}}$ & 68.4 & 21.1 & 10.4 & 72.7 & 10.7 & 16.7 & 75.3 & 12.2 & 12.4 \\
BC-3$_{\text{Q}}$ & 65.3 & 32.9 &  1.8 & 82.7 & 12.2 &  5.1 & 79.9 & 16.9 &  3.2 \\
\midrule
BC-5$_{\text{G}}$ & 70.4 & 22.4 &  7.1 & 73.8 & 11.2 & 15.0 & 76.6 & 10.1 & 13.3 \\
BC-5$_{\text{Q}}$ & 68.3 & 30.7 &  1.0 & 82.2 & 13.6 &  4.2 & 81.9 & 15.1 &  3.0 \\
\bottomrule
\end{tabular}}
\subcaption{Moving threats.}
\label{tab:imggraph-moving}
\end{minipage}
\caption{\textit{\textbf{Image + Graph}} modality results, grouped by map difficulty. Fail = \hF, Save = \hS, Timeout = \hT. Subscripts denote the operator/civilian model: G = Gemma, Q = Qwen.}
\label{tab:imggraph-combined}
\end{table*}

\section{GPT-5.4 Results}
\label{app:gpt-results}

We additionally evaluated \texttt{GPT-5.4} to test whether our findings extend to stronger closed models. The model was used as both operator and civilian under the \textit{\textbf{Image}} setup with static threats, across all five communication strategies. Due to its substantial monetary cost, we restricted the evaluation to Easy 1 and Hard 1, which we consider the most representative maps of their respective tiers. Running these two maps alone cost approximately \$400 (about \$200 per map for all five communication strategies); extending the evaluation to the full grid of nine maps, three input modalities, and two threat settings would have raised the cost to roughly \$10{,}800.

Results are reported in Table~\ref{tab:gpt-results}. With a stronger model, Save rates increase markedly: above 80\% on Easy 1 and above 50\% on Hard 1, well beyond the best open-weight result (41.7\% for Gemma on Hard 1). Fail rates under Narrowcast are close to zero on both maps, indicating that GPT-5.4 reliably avoids guiding civilians into threats: on Easy 1 this translates into very high Save rates, while on Hard 1 unresolved cases are absorbed by Timeouts (40.0\% for NC-C) rather than Fails (2.3\%). From a safety standpoint this is highly desirable: virtually no civilian comes into contact with a threat, even on the hardest map.

A notable shift relative to the open-weight setup is the relative ranking of NC-C and NC-D. With Gemma and Qwen, NC-D consistently underperformed NC-C on Save rates because of looping-induced Timeouts. With \texttt{GPT-5.4} this trend reverses: NC-D outperforms NC-C across all outcome columns on both maps, suggesting that the stronger reasoning capability of the model is better suited to exploit the structured threat-and-safe-zone preamble required by NC-D.

Broadcast strategies also benefit from the stronger model, but their Save rates remain below those of Narrowcast, with NC-D maintaining the largest margin (e.g., on Hard 1, NC-D Save $= 65.6\%$ vs.\ best BC Save $= 54.6\%$).

\begin{table}[ht]
\centering
\footnotesize
\setlength{\tabcolsep}{3pt}
\begin{tabular}{l
  >{\columncolor{gray!20}}c >{\columncolor{gray!20}}c >{\columncolor{gray!20}}c  
  ccc
  }
\toprule
\multirow{2}{*}{\hStaticTable}
  & \multicolumn{3}{>{\columncolor{gray!20}}c}{Easy}
  & \multicolumn{3}{c}{Hard} \\
\cmidrule(lr){2-4} \cmidrule(lr){5-7}
& \hF & \hS & \hT & \hF & \hS & \hT  \\
\midrule
NC-C  &  2.7 & 79.0 & 18.3 &  2.3 & 57.7 & 40.0 \\
NC-D &  3.0 & 80.7 & 16.3 &  3.7 & 65.6 & 30.7 \\
\midrule
BC-1 & 10.7 & 74.6 & 14.7 & 13.7 & 54.3 & 32.0 \\
BC-3 & 10.7 & 75.3 & 14.0 & 16.7 & 53.0 & 30.3 \\
BC-5 & 12.3 & 75.7 & 12.0 & 13.9 & 54.6 & 31.5 \\
\bottomrule
\end{tabular}
\caption{GPT-5.4 results on Easy 1 and Hard 1 maps \textit{\textbf{Image}} setup.}
\label{tab:gpt-results}
\end{table}

\section{Experiments Full Details}
\label{app:experiments-full-details}

\paragraph{Model Serving.} Both VLMs (\texttt{Qwen3-VL-30B} \texttt{-FP8} and \texttt{Gemma-3-27B}) are served through vLLM Docker containers, each allocated two NVIDIA A40 GPUs with tensor parallelism. Decoding parameters are identical across the two models: top-$p = 0.8$ and top-$k = 20$. Temperature is set to $0.7$ for the operator and $0.2$ for civilians.

\paragraph{Cross-model evaluation.} In the main text we reported results obtained by pairing the same model as both operator and civilian. To rule out the possibility that this configuration inflates performance through implicit alignment between the operator's outputs and the civilian's interpretation, we ran additional experiments in which operator and civilian are different models. We evaluated all four operator/civilian pairings of Gemma and Qwen (subscripts G and Q in the tables) across all five communication strategies, both threat settings, and all three input modalities. Running two distinct VLMs in parallel required holding two jobs simultaneously on a shared GPU cluster, which made it impractical to scale this evaluation to the full set of nine maps. We therefore restricted the cross-model setting to the first map of each difficulty tier (Easy 1, Medium 1, Hard 1).

Tables~\ref{tab:cross-model-combined}
and~\ref{tab:cross-model-modality-combined} report the results. The trends match those observed in the same-model setting. Narrowcast strategies achieve lower Fail rates than Broadcast across all map difficulties and threat settings, and the relative ranking between Gemma and Qwen operators is consistent with the main text. The operator model also dominates the outcome over the civilian model, an effect most evident in Narrowcast and weaker in Broadcast. Overall, the cross-model results confirm that the trends in the main text are not artifacts of same-model pairing.

\begin{table*}[ht]
\centering
\footnotesize
\setlength{\tabcolsep}{3pt}
\begin{minipage}{0.49\textwidth}
\centering
\resizebox{\linewidth}{!}{%
\begin{tabular}{l
  >{\columncolor{gray!20}}c >{\columncolor{gray!20}}c >{\columncolor{gray!20}}c
  ccc
  >{\columncolor{gray!20}}c >{\columncolor{gray!20}}c >{\columncolor{gray!20}}c
  }
\toprule
\multirow{2}{*}{\hStaticTable}
  & \multicolumn{3}{>{\columncolor{gray!20}}c}{Easy}
  & \multicolumn{3}{c}{Medium}
  & \multicolumn{3}{>{\columncolor{gray!20}}c}{Hard} \\
\cmidrule(lr){2-4} \cmidrule(lr){5-7} \cmidrule(lr){8-10}
& \hF & \hS & \hT & \hF & \hS & \hT & \hF & \hS & \hT \\
\midrule
  NC-C$_{\text{GG}}$  &  11.0 & 64.7 & 24.3 & 11.0 & 46.7 & 42.3 &  9.7 & 41.7 & 48.6 \\
  NC-C$_{\text{GQ}}$  &   8.7 & 64.3 & 27.0 & 12.0 & 44.7 & 43.3 & 10.3 & 41.7 & 48.0 \\
  NC-C$_{\text{QQ}}$  &  10.0 & 47.0 & 43.0 &  3.0 & 31.7 & 65.3 & 13.7 & 30.0 & 56.3 \\
  NC-C$_{\text{QG}}$  &   9.7 & 49.3 & 41.0 &  4.0 & 31.3 & 64.7 & 16.3 & 24.0 & 59.7 \\
\midrule
  NC-D$_{\text{GG}}$ &   6.3 & 58.7 & 35.0 &  6.7 & 41.7 & 51.6 &  5.7 & 30.7 & 63.6 \\
  NC-D$_{\text{GQ}}$ &   6.0 & 59.3 & 34.7 &  6.0 & 40.0 & 54.0 &  7.0 & 27.3 & 65.7 \\
  NC-D$_{\text{QQ}}$ &  10.7 & 37.0 & 52.3 &  6.3 & 26.7 & 67.0 & 14.0 & 23.3 & 62.7 \\
  NC-D$_{\text{QG}}$ &   7.7 & 42.3 & 50.0 &  6.0 & 21.3 & 72.7 &  8.0 & 21.3 & 70.7 \\
\midrule
  BC-1$_{\text{GG}}$ &  40.7 & 37.0 & 22.3 & 23.7 & 20.0 & 56.3 & 38.7 & 26.3 & 35.0 \\
  BC-1$_{\text{GQ}}$ &  47.3 & 42.7 & 10.0 & 27.7 & 18.3 & 54.0 & 56.7 & 12.3 & 31.0 \\
  BC-1$_{\text{QQ}}$ &  36.7 & 49.6 & 13.7 & 34.7 & 23.3 & 42.0 & 46.7 & 30.3 & 23.0 \\
  BC-1$_{\text{QG}}$ &  30.3 & 37.0 & 32.7 & 28.3 & 23.3 & 48.4 & 42.6 & 33.7 & 23.7 \\
\midrule
  BC-3$_{\text{GG}}$ &  36.0 & 36.3 & 27.7 & 24.7 & 20.0 & 55.3 & 39.0 & 19.7 & 41.3 \\
  BC-3$_{\text{GQ}}$ &  45.3 & 44.7 & 10.0 & 36.0 & 18.0 & 46.0 & 58.7 & 13.0 & 28.3 \\
  BC-3$_{\text{QQ}}$ &  30.3 & 51.7 & 18.0 & 39.3 & 21.7 & 39.0 & 43.4 & 33.3 & 23.3 \\
  BC-3$_{\text{QG}}$ &  36.7 & 38.0 & 25.3 & 26.7 & 19.7 & 53.6 & 43.6 & 31.7 & 24.7 \\
\midrule
  BC-5$_{\text{GG}}$ &  38.0 & 35.3 & 26.7 & 21.7 & 22.3 & 56.0 & 44.0 & 23.3 & 32.7 \\
  BC-5$_{\text{GQ}}$ &  41.0 & 48.0 & 11.0 & 34.0 & 17.0 & 49.0 & 57.4 & 15.3 & 27.3 \\
  BC-5$_{\text{QQ}}$ &  33.3 & 52.0 & 14.7 & 35.0 & 22.3 & 42.7 & 50.3 & 32.0 & 17.7 \\
  BC-5$_{\text{QG}}$ &  28.7 & 39.3 & 32.0 & 25.3 & 23.7 & 51.0 & 47.0 & 31.3 & 21.7 \\
\bottomrule
\end{tabular}}
\subcaption{Static threats.}
\label{tab:cross-model-left}
\end{minipage}%
\hfill
\begin{minipage}{0.49\textwidth}
\centering
\resizebox{\linewidth}{!}{%
\begin{tabular}{l
  >{\columncolor{gray!20}}c >{\columncolor{gray!20}}c >{\columncolor{gray!20}}c
  ccc
  >{\columncolor{gray!20}}c >{\columncolor{gray!20}}c >{\columncolor{gray!20}}c
  }
\toprule
\multirow{2}{*}{\hMovingTable}
  & \multicolumn{3}{>{\columncolor{gray!20}}c}{Easy}
  & \multicolumn{3}{c}{Medium}
  & \multicolumn{3}{>{\columncolor{gray!20}}c}{Hard} \\
\cmidrule(lr){2-4} \cmidrule(lr){5-7} \cmidrule(lr){8-10}
& \hF & \hS & \hT & \hF & \hS & \hT & \hF & \hS & \hT \\
\midrule
  NC$_{\text{GG}}$   & 40.0 & 47.7 & 12.3 & 50.7 & 26.3 & 23.0 & 49.3 & 20.7 & 30.0 \\
  NC$_{\text{GQ}}$   & 39.7 & 45.7 & 14.7 & 49.3 & 17.3 & 33.3 & 57.0 & 21.3 & 21.7 \\
  NC$_{\text{QQ}}$   & 43.3 & 35.0 & 21.7 & 41.0 & 17.3 & 41.7 & 52.0 & 17.3 & 30.7 \\
  NC$_{\text{QG}}$   & 38.3 & 34.7 & 27.0 & 41.7 & 16.3 & 42.0 & 54.3 & 15.3 & 30.3 \\
\midrule
  NCB$_{\text{GG}}$  & 27.7 & 48.3 & 24.0 & 40.3 & 19.0 & 40.7 & 43.7 & 20.7 & 35.7 \\
  NCB$_{\text{GQ}}$  & 34.0 & 40.7 & 25.3 & 38.7 & 15.7 & 45.7 & 42.0 & 15.7 & 42.3 \\
  NCB$_{\text{QQ}}$  & 48.0 & 25.7 & 26.3 & 37.7 & 13.7 & 48.7 & 44.7 & 15.7 & 39.7 \\
  NCB$_{\text{QG}}$  & 45.0 & 29.0 & 26.0 & 37.7 & 12.0 & 50.3 & 47.0 & 15.7 & 37.3 \\
\midrule
  BC-1$_{\text{GG}}$ & 51.0 & 34.3 & 14.7 & 57.3 & 13.0 & 29.7 & 61.7 & 18.0 & 20.3 \\
  BC-1$_{\text{GQ}}$ & 65.0 & 29.7 &  5.3 & 70.7 & 10.3 & 19.0 & 86.0 &  5.7 &  8.3 \\
  BC-1$_{\text{QQ}}$ & 59.0 & 34.3 &  6.7 & 79.7 &  9.3 & 11.0 & 73.0 & 19.0 &  8.0 \\
  BC-1$_{\text{QG}}$ & 58.0 & 27.0 & 15.0 & 57.7 & 12.0 & 30.3 & 67.7 & 18.7 & 13.7 \\
\midrule
  BC-3$_{\text{GG}}$ & 57.7 & 32.7 &  9.7 & 61.3 & 10.3 & 28.3 & 71.3 & 13.7 & 15.0 \\
  BC-3$_{\text{GQ}}$ & 63.0 & 30.3 &  6.7 & 77.7 & 11.3 & 11.0 & 81.3 &  7.0 & 11.7 \\
  BC-3$_{\text{QQ}}$ & 60.3 & 33.7 &  6.0 & 80.3 & 11.3 &  8.3 & 78.3 & 18.0 &  3.7 \\
  BC-3$_{\text{QG}}$ & 58.3 & 29.7 & 12.0 & 62.7 & 14.0 & 23.3 & 75.0 & 16.3 &  8.7 \\
\midrule
  BC-5$_{\text{GG}}$ & 62.3 & 26.7 & 11.0 & 66.3 & 11.3 & 22.3 & 66.0 & 15.7 & 18.3 \\
  BC-5$_{\text{GQ}}$ & 65.7 & 29.0 &  5.3 & 78.7 & 10.7 & 10.7 & 82.0 &  6.0 & 12.0 \\
  BC-5$_{\text{QQ}}$ & 59.0 & 36.7 &  4.3 & 75.3 & 14.0 & 10.7 & 80.3 & 12.7 &  7.0 \\
  BC-5$_{\text{QG}}$ & 57.7 & 31.3 & 11.0 & 70.7 & 11.7 & 17.7 & 68.7 & 22.7 &  8.7 \\
\bottomrule
\end{tabular}}
\subcaption{Moving threats.}
\label{tab:cross-model-right}
\end{minipage}
\caption{Cross-model results grouped by map difficulty. Subscripts denote operator/civilian model pairing: GG = Gemma/Gemma, QQ = Qwen/Qwen, GQ = Gemma operator + Qwen civilians, QG = Qwen operator + Gemma civilians.}
\label{tab:cross-model-combined}
\end{table*}

\begin{table*}[ht]
\centering
\footnotesize
\setlength{\tabcolsep}{3pt}
\begin{minipage}{0.49\textwidth}
\centering
\resizebox{\linewidth}{!}{%
\begin{tabular}{l
  >{\columncolor{gray!20}}c >{\columncolor{gray!20}}c >{\columncolor{gray!20}}c
  ccc
  >{\columncolor{gray!20}}c >{\columncolor{gray!20}}c >{\columncolor{gray!20}}c
  }
\toprule
\multirow{2}{*}{\hStaticTable}
  & \multicolumn{3}{>{\columncolor{gray!20}}c}{Graph}
  & \multicolumn{3}{c}{Image}
  & \multicolumn{3}{>{\columncolor{gray!20}}c}{image + Graph} \\
\cmidrule(lr){2-4} \cmidrule(lr){5-7} \cmidrule(lr){8-10}
& \hF & \hS & \hT & \hF & \hS & \hT & \hF & \hS & \hT \\
\midrule
  NC-C$_{\text{GG}}$ & 32.1 & 30.1 & 37.8 & 10.6 & 51.0 & 38.4 & 11.9 & 28.2 & 59.9 \\
  NC-C$_{\text{GQ}}$ & 31.3 & 30.2 & 38.4 & 10.3 & 50.2 & 39.4 & 12.1 & 25.2 & 62.7 \\
  NC-C$_{\text{QQ}}$ & 40.4 & 35.7 & 23.9 &  8.9 & 36.2 & 54.9 & 38.9 & 38.7 & 22.4 \\
  NC-C$_{\text{QG}}$ & 39.3 & 34.3 & 26.3 & 10.0 & 34.9 & 55.1 & 37.6 & 40.0 & 22.4 \\
\midrule
  NC-D$_{\text{GG}}$ &  9.8 & 21.3 & 68.9 &  6.2 & 43.7 & 50.1 &  4.2 & 21.9 & 73.9 \\
  NC-D$_{\text{GQ}}$ & 11.7 & 20.9 & 67.4 &  6.3 & 42.2 & 51.4 &  4.8 & 18.9 & 76.3 \\
  NC-D$_{\text{QQ}}$ & 25.4 & 36.6 & 38.0 & 10.3 & 29.0 & 60.7 & 19.8 & 40.9 & 39.3 \\
  NC-D$_{\text{QG}}$ & 18.8 & 32.1 & 49.1 &  7.2 & 28.3 & 64.4 & 17.0 & 37.6 & 45.4 \\
\midrule
  BC-1$_{\text{GG}}$ & 49.6 & 16.6 & 33.9 & 34.3 & 27.8 & 37.9 & 36.1 & 26.6 & 37.3 \\
  BC-1$_{\text{GQ}}$ & 57.8 & 16.8 & 25.4 & 43.9 & 24.4 & 31.7 & 47.8 & 24.4 & 27.8 \\
  BC-1$_{\text{QQ}}$ & 56.8 & 20.3 & 22.9 & 39.3 & 34.4 & 26.2 & 50.7 & 28.9 & 20.4 \\
  BC-1$_{\text{QG}}$ & 40.9 & 16.4 & 42.7 & 33.8 & 31.3 & 34.9 & 39.0 & 25.8 & 35.2 \\
\midrule
  BC-3$_{\text{GG}}$ & 48.4 & 16.4 & 35.1 & 33.2 & 25.3 & 41.4 & 36.1 & 24.9 & 39.0 \\
  BC-3$_{\text{GQ}}$ & 56.7 & 17.7 & 25.7 & 46.7 & 25.2 & 28.1 & 47.1 & 24.4 & 28.4 \\
  BC-3$_{\text{QQ}}$ & 59.1 & 20.0 & 20.9 & 37.7 & 35.6 & 26.8 & 50.2 & 29.0 & 20.8 \\
  BC-3$_{\text{QG}}$ & 42.4 & 18.2 & 39.3 & 35.7 & 29.8 & 34.6 & 39.6 & 25.7 & 34.8 \\
\midrule
  BC-5$_{\text{GG}}$ & 44.4 & 18.4 & 37.1 & 34.6 & 27.0 & 38.4 & 36.0 & 24.6 & 39.4 \\
  BC-5$_{\text{GQ}}$ & 56.9 & 16.9 & 26.2 & 44.1 & 26.8 & 29.1 & 45.6 & 22.9 & 31.6 \\
  BC-5$_{\text{QQ}}$ & 54.8 & 21.3 & 23.9 & 39.6 & 35.4 & 25.0 & 52.6 & 27.0 & 20.4 \\
  BC-5$_{\text{QG}}$ & 44.3 & 20.8 & 34.9 & 33.7 & 31.4 & 34.9 & 39.3 & 24.6 & 36.1 \\
\bottomrule
\end{tabular}}
\subcaption{Static threats.}
\label{tab:cross-model-modality-left}
\end{minipage}%
\hfill
\begin{minipage}{0.49\textwidth}
\centering
\resizebox{\linewidth}{!}{%
\begin{tabular}{l
  >{\columncolor{gray!20}}c >{\columncolor{gray!20}}c >{\columncolor{gray!20}}c
  ccc
  >{\columncolor{gray!20}}c >{\columncolor{gray!20}}c >{\columncolor{gray!20}}c
  }
\toprule
\multirow{2}{*}{\hMovingTable}
  & \multicolumn{3}{>{\columncolor{gray!20}}c}{Graph}
  & \multicolumn{3}{c}{Image}
  & \multicolumn{3}{>{\columncolor{gray!20}}c}{image + Graph} \\
\cmidrule(lr){2-4} \cmidrule(lr){5-7} \cmidrule(lr){8-10}
& \hF & \hS & \hT & \hF & \hS & \hT & \hF & \hS & \hT \\
\midrule
  NC$_{\text{GG}}$   & 61.7 & 19.3 & 19.0 & 46.7 & 31.6 & 21.8 & 56.1 & 17.6 & 26.3 \\
  NC$_{\text{GQ}}$   & 61.8 & 19.6 & 18.7 & 48.7 & 28.1 & 23.2 & 53.6 & 18.3 & 28.1 \\
  NC$_{\text{QQ}}$   & 63.2 & 25.3 & 11.4 & 45.4 & 23.2 & 31.3 & 62.0 & 27.9 & 10.1 \\
  NC$_{\text{QG}}$   & 58.8 & 25.8 & 15.4 & 44.8 & 22.1 & 33.1 & 60.2 & 31.1 &  8.7 \\
\midrule
  NCB$_{\text{GG}}$  & 52.3 & 12.3 & 35.3 & 37.2 & 29.3 & 33.4 & 41.9 & 15.2 & 42.9 \\
  NCB$_{\text{GQ}}$  & 50.0 & 14.0 & 36.0 & 38.2 & 24.0 & 37.8 & 45.0 & 13.2 & 41.8 \\
  NCB$_{\text{QQ}}$  & 59.1 & 22.1 & 18.8 & 43.4 & 18.3 & 38.2 & 54.0 & 27.0 & 19.0 \\
  NCB$_{\text{QG}}$  & 55.1 & 22.2 & 22.7 & 43.2 & 18.9 & 37.9 & 48.2 & 29.1 & 22.7 \\
\midrule
  BC-1$_{\text{GG}}$ & 73.8 & 11.0 & 15.2 & 56.7 & 21.8 & 21.6 & 61.9 & 19.0 & 19.1 \\
  BC-1$_{\text{GQ}}$ & 78.6 & 13.3 &  8.1 & 73.9 & 15.2 & 10.9 & 71.8 & 18.2 & 10.0 \\
  BC-1$_{\text{QQ}}$ & 76.6 & 16.2 &  7.2 & 70.6 & 20.9 &  8.6 & 70.4 & 21.6 &  8.0 \\
  BC-1$_{\text{QG}}$ & 66.2 & 15.4 & 18.3 & 61.1 & 19.2 & 19.7 & 64.6 & 19.4 & 16.0 \\
\midrule
  BC-3$_{\text{GG}}$ & 72.3 & 10.7 & 17.0 & 63.4 & 18.9 & 17.7 & 65.8 & 16.8 & 17.4 \\
  BC-3$_{\text{GQ}}$ & 80.7 & 10.9 &  8.4 & 74.0 & 16.2 &  9.8 & 74.3 & 15.1 & 10.6 \\
  BC-3$_{\text{QQ}}$ & 76.4 & 16.7 &  6.9 & 73.0 & 21.0 &  6.0 & 71.7 & 21.1 &  7.2 \\
  BC-3$_{\text{QG}}$ & 69.9 & 14.6 & 15.6 & 65.3 & 20.0 & 14.7 & 66.6 & 19.4 & 14.0 \\
\midrule
  BC-5$_{\text{GG}}$ & 76.0 &  9.3 & 14.7 & 64.9 & 17.9 & 17.2 & 70.2 & 15.8 & 14.0 \\
  BC-5$_{\text{GQ}}$ & 78.7 & 11.6 &  9.8 & 75.4 & 15.2 &  9.3 & 73.9 & 16.9 &  9.2 \\
  BC-5$_{\text{QQ}}$ & 76.8 & 15.2 &  8.0 & 71.6 & 21.1 &  7.3 & 73.1 & 21.7 &  5.2 \\
  BC-5$_{\text{QG}}$ & 68.1 & 15.2 & 16.7 & 65.7 & 21.9 & 12.4 & 67.4 & 19.2 & 13.3 \\
\bottomrule
\end{tabular}}
\subcaption{Moving threats.}
\label{tab:cross-model-modality-right}
\end{minipage}
\caption{Cross-model modality interaction results averaged across all maps. Subscripts denote operator/civilian model pairing: GG = Gemma/Gemma, QQ = Qwen/Qwen, GQ = Gemma/Qwen, QG = Qwen/Gemma.}
\label{tab:cross-model-modality-combined}
\end{table*}

\end{document}